\documentclass[journal]{IEEEtran}

\usepackage{graphicx}
\usepackage{tabularx}
\usepackage[mathscr]{euscript}
\usepackage{amsmath}
\usepackage{multirow}
\usepackage{amsfonts}
\usepackage{float}
\usepackage{soul}

\newcolumntype{x}[1]{>{\centering\arraybackslash}p{#1}}
\newcolumntype{Y}{>{\centering\arraybackslash}X}
\usepackage{balance}
\usepackage{color}
\usepackage{threeparttable}
\newcommand{\textBF}[1]{%
    \pdfliteral direct {2 Tr 0.3 w} %the second factor is the boldness
     #1%
    \pdfliteral direct {0 Tr 0 w}%
}

\ifCLASSINFOpdf
\else
\fi

\usepackage{amsmath}
\usepackage{amssymb}% http://ctan.org/pkg/amssymb
\usepackage{pifont}% http://ctan.org/pkg/pifont
\usepackage{subcaption}
\usepackage{comment}
\usepackage{hyperref}
\usepackage{booktabs} % For formal tables
\usepackage{mathrsfs}
\usepackage{amsfonts}
\usepackage{multirow}
\usepackage{balance}

\begin{document}

\title{TraverseNet: Unifying Space and Time in Message Passing for Traffic Forecasting}

\author{Zonghan Wu, Da Zheng, Shirui Pan,
Quan Gan, 
Guodong Long,
George Karypis~\IEEEmembership{Fellow,~IEEE}

\IEEEcompsocitemizethanks{
	\IEEEcompsocthanksitem Z. Wu,  G. Long are with Centre for Artificial Intelligence, FEIT, University of Technology Sydney, NSW 2007, Australia (E-mail: zonghan.wu-3@student.uts.edu.au;  guodong.long@uts.edu.au;).
	\IEEEcompsocthanksitem D. Zheng, Q. Gan, G. Karypis are with Amazon (E-mail: dzzhen@amazon.com; quagan@amazon.com;
	gkarypis@amazon.com)
	\IEEEcompsocthanksitem S. Pan is with Monash University. From August 2022, he will be with the School of Information and Communication Technology, Griffith University, Southport, QLD 4222, Australia (Email: shiruipan@ieee.org). 
	\IEEEcompsocthanksitem \textit{Corresponding Authors: Da Zheng and Shirui Pan.}
}

\thanks{Manuscript received Dec xx, 2018; revised Dec xx, 201x. This research was supported by an ARC Future Fellowship  (FT210100097).}}

\markboth{Journal of \LaTeX\ Class Files,~Vol.~xx, No.~xx, August~2019}%
{Shell \MakeLowercase{\textit{et al.}}: Bare Demo of IEEEtran.cls for IEEE Journals}

\maketitle

\begin{abstract}
This paper aims to unify spatial dependency and temporal dependency in a non-Euclidean space while capturing the inner spatial-temporal dependencies for \textcolor{black}{traffic data}. For spatial-temporal attribute entities with topological structure, the space-time is consecutive and unified while each node's current status is influenced by its neighbors' past states over variant periods of each neighbor. Most spatial-temporal neural networks \textcolor{black}{for traffic forecasting} study spatial dependency and temporal correlation separately in processing, gravely impaired the spatial-temporal integrity, and ignore the fact that the neighbors' temporal dependency period for a node can be delayed and dynamic. To model this actual condition, we propose TraverseNet, a novel spatial-temporal graph neural network, viewing space and time as an inseparable whole, to mine spatial-temporal graphs while exploiting the evolving spatial-temporal dependencies for each node via message traverse mechanisms. Experiments with ablation and parameter studies have validated the effectiveness of the proposed TraverseNet, and the detailed implementation can be found from \href{https://github.com/nnzhan/TraverseNet}{https://github.com/nnzhan/TraverseNet}.
\end{abstract}


\IEEEpeerreviewmaketitle

\section{Introduction}
% The rise of graph neural networks has facilitated researches in a wide range of  graph-related tasks including node classification \cite{defferrard2016convolutional, kipf2017semi,wu2021beyond}, graph classification \cite{xu2019how, ying2018hierarchical}, anomaly detection \cite{liu2021anomaly, liu2021anomaly2}, %graph generation \cite{you2018graphrnn, bojchevski2018netgan}, 
% spatial-temporal graph forecasting \cite{li2018diffusion, yu2018spatio}.
% Among them, spatial-temporal graph forecasting is a new research topic emerged very recently. It is widely applied to solve real world problems such as EEG signals recognition \cite{zhang2019graph}, skeleton-based action recognition \cite{yan2018spatial}, and traffic forecasting \cite{pan2019urban}.
\textcolor{black}{Spatial-temporal graph neural networks for traffic forecasting is a new research topic \textcolor{black}{that} emerged very recently.}
It involves message passing in a dynamic system in which node features are constantly changing over time. Spatial-temporal graph neural networks assume that the nodes in a topological structure contain spatial-temporal dependencies where a node's future pattern is subject to its neighbors' historical results as well as its own past records. 
Specifically, the traffic conditions on a particular road at a given time depend not only on that road's previous traffic conditions but also on the traffic conditions of its adjacent roads several minutes ago, as it takes time for vehicles to travel from one location to the next. Hence, how to effectively exploit and preserve both spatial and temporal inner-dependency turns into an essential challenge to answer.
\begin{figure}
        \begin{subfigure}{0.23\textwidth}   
            \centering \includegraphics[width=\textwidth]{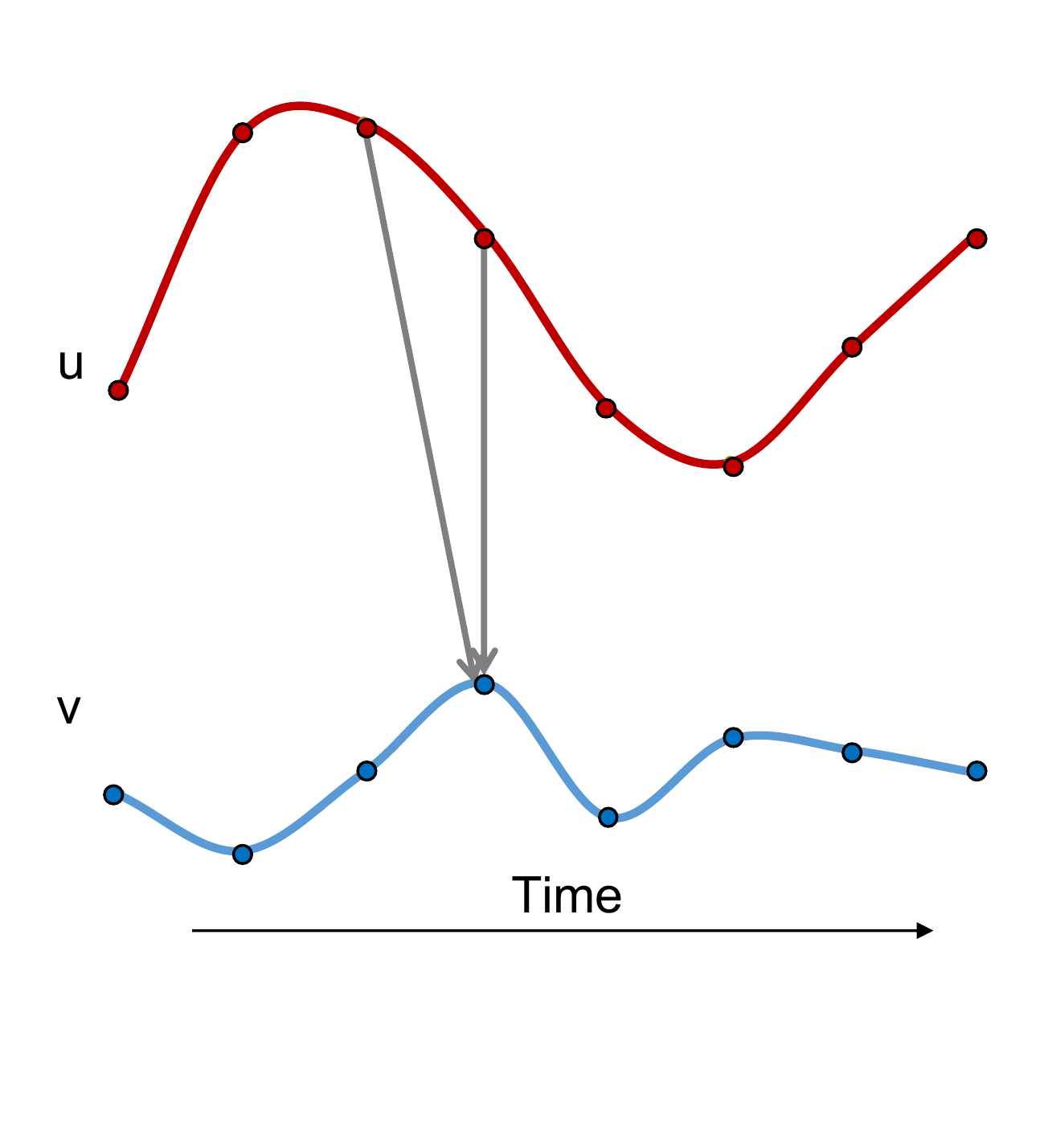}
            \caption[RNN-based Methods]%
            {{\footnotesize\rm RNN-based Methods. The node $v$ receives raw information of its neighbor $u$ at the current step and latent information from the last time step. }}    
            \label{fig:rnn}
        \end{subfigure}
        \hfill
        \begin{subfigure}{0.23\textwidth}   
            \centering 
            \includegraphics[width=\textwidth]{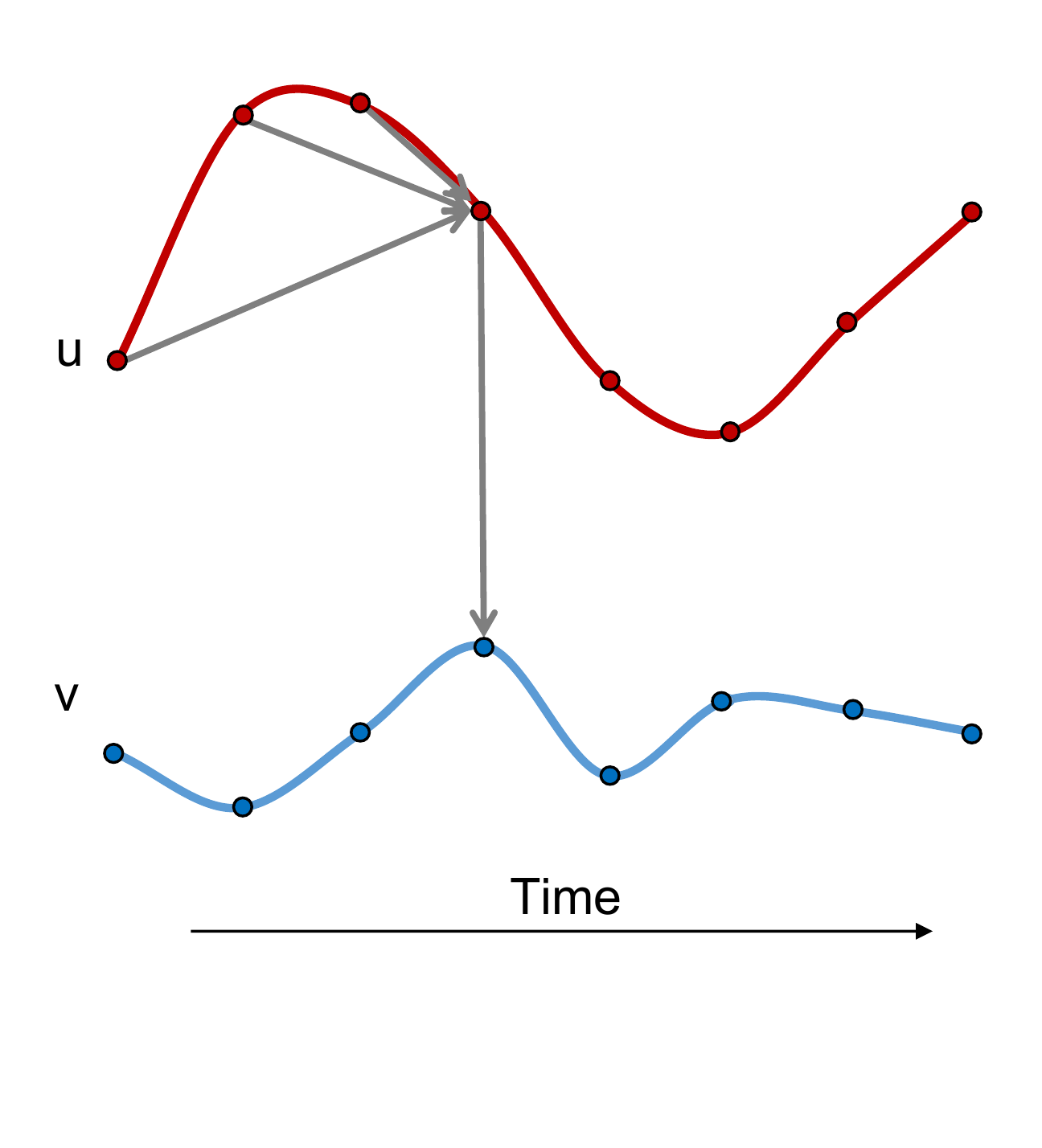}
            \caption[CNN-based Methods]%
            {{\rm\footnotesize CNN-based Methods. The node $v$ receives information from its neighbor $u$ at the current time step and a short window of recent time steps.}}    
            \label{fig:cnn}
        \end{subfigure}
        \\[5pt]
        \begin{subfigure}{0.23\textwidth}
            \centering 
            \includegraphics[width=\textwidth]{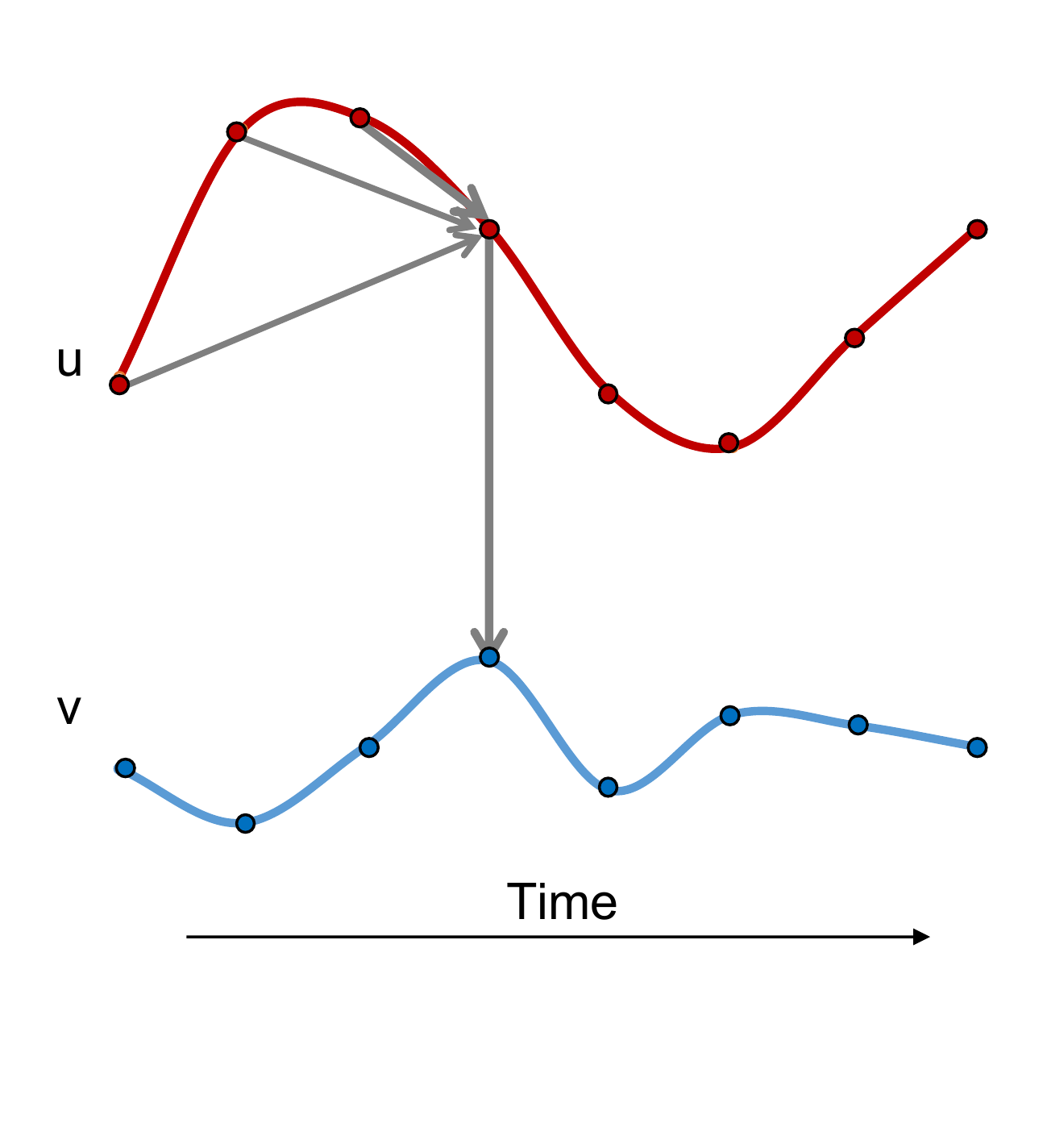}
            \caption[Attention-based Methods]%
            {{\rm\footnotesize Attention-based Methods. The node $v$ receives information from its neighbor $u$ at the current time step and a conditional weighted sum of information from recent time steps based on the states between two sides. }}    
            \label{fig:atten}
        \end{subfigure}
        \hfill
        \begin{subfigure}{0.23\textwidth}   
            \centering 
            \includegraphics[width=\textwidth]{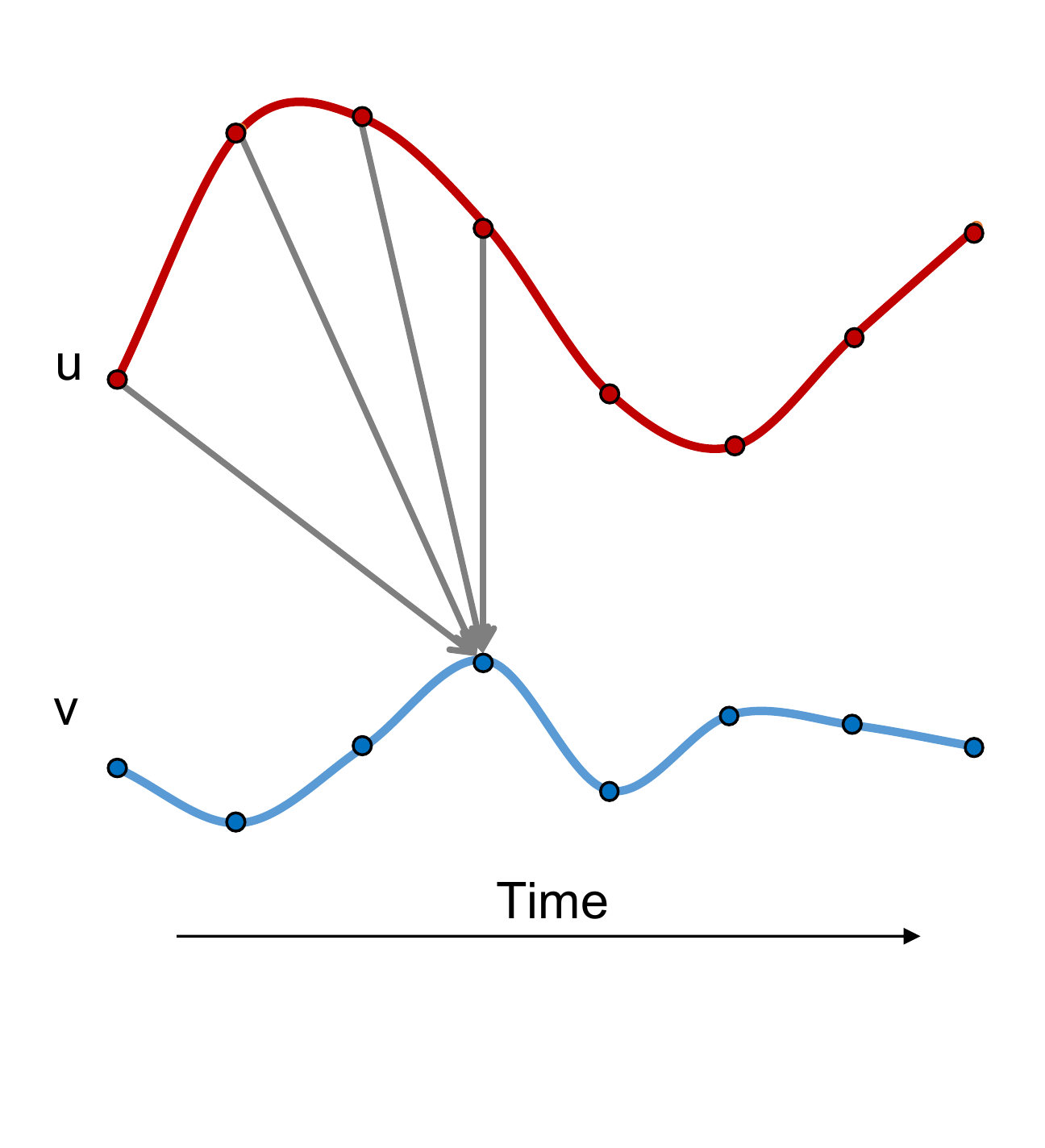}
            \caption[Our method]%
            {{\rm\footnotesize Our method. The node $v$ selectively receives information of its neighbor $u$ within a period of time directly. The importance weight of each time step is condition on the corresponding states between two sides.  }}    
            \label{fig:our}
        \end{subfigure}
        \caption[Message Passing Diagrams]
        {\small Message Passing Diagrams of Different STGNNs. The blue line denotes the evolving node features of a node $v$. The red line is the evolving node features of its neighbor $u$. The arrows are used to illustrate the message passing paradigms in different spatial-temporal approaches.} 
        \label{fig:diag}
\end{figure}

Existing graph neural networks or graph convolutional neural networks can not solve this problem alone. The main reason is \textcolor{black}{that} they are designed to handle static node features so that only spatial dependencies can be captured. To jointly capture spatial topology and temporal sequence in a spatial-temporal graph (ST-Graph), researchers naturally considered combining graph convolutional neural networks (GCNs) and recurrent neural networks (RNNs) \cite{seo2018structured, li2018diffusion, zhang2018gaan}. While effective in fusing topological information into temporal sequence learning, RNN-based frameworks are inefficient in capturing long-range spatial-temporal dependencies. As illustrated \textcolor{black}{in} Figure \ref{fig:rnn}, during each recurrent step, they only allow a node to be aware of its neighbors' current inputs and its neighbors' previous hidden states. Information loss is inevitable when processing long sequences.

Meanwhile, capitalizing on parallel computing and stable convolutional propagation, CNN-based ST-Graph frameworks  have received considerable attention~\cite{yan2018spatial, yu2018spatio}. They stack temporal convolution layers and graph convolution layers to capture local spatial-temporal dependencies.  As illustrated in Figure \ref{fig:cnn}, they first use a small 1D convolutional kernel to propagate \textcolor{black}{nodes'} temporal information to the current time step, then use the graph convolution to pass the aggregated temporal information of a node's neighbors to the node itself. To capture the long-range spatial-temporal dependencies, they face the trade-off between kernel size and number of layers. If a large 1D kernel is used to retain long-range spatial-temporal relations, the model has to be shallow because of a small number of layers. Alternatively, if a small kernel is used, a large number of 1D CNN layers and graph convolutional layers are requested, resulting in efficiency issues.

%However, in order to persist long-range temporal memory, a large number of 1D CNN layers are requested, resulting in the linearly growing receptive field size, leading the efficiency issue back to original point. Moreover, the fixed temporal recall weights essentially limit both RNN-based and CNN-based ST-Graph frameworks to nimbly explore and persist the useful memory from variant historical stamps. In the meantime, information loss is inevitable when the persisted memory transmit to the current state from the past.
The attention mechanism is known for its efficiency in delivering important information \cite{vaswani2017attention,tang2018self}. A number of studies integrate spatial attention with temporal attention for spatial-temporal data modeling \cite{guo2019attention,park2020stgrat,wang2020traffic}. The information flow diagram of attention-based methods is similar to that of CNN-based methods. Instead of using convolutional kernels, they apply attention mechanisms for information aggregation. As illustrated \textcolor{black}{in} Figure \ref{fig:atten}, they first use temporal attention to pass important historical information of each node to its current step. Then they aggregate neighborhood information selectively for each node by spatial attention. In this way, each node will receive its neighbor's temporal information indirectly. However, a node cannot determine which period of temporal information from the neighborhood is more relevant to itself because the computation of a neighbor's temporal representation is independent of the central node itself.

To summarize, existing approaches either model spatial-temporal dependencies locally or model spatial correlations and temporal correlations separately. They prevent a node from being directly aware of its \textcolor{black}{neighborhood's} long-range historical information.  
In fact, a node’s current state may depend on its neighbors' previous states within a certain period of time. As illustrated \textcolor{black}{in} Figure \ref{fig:our}, the rise of a node's curve may exert influence on its neighbors several time steps later because of physical distance. For example, a traffic congestion of a road will cause another congestion of its nearby roads 15 minutes later. %We call this phenomenon space-induced temporal delay. 
It suggests that treating spatial correlations and temporal correlations locally or separately is inappropriate. \textcolor{black}{In fact, the spatial-temporal dependency is a whole which can not be separated into the spatial dependency and the temporal dependency (We refer this as spatial-temporal integrity).} Additionally, existing ST-GNNs tend to simply stack different layers (e.g., inception layer, dilated convolutional layer and RNN/CNN layer) together, resulting in overly complex and cumbersome architecture. Such construction confuses the significance and contribution of each kind of layer. 

To overcome the above challenges, we present TraverseNet, a novel spatial-temporal neural network for structured data. TraverseNet processes a spatial-temporal graph as an inseparable entity.  Our specially designed message traverse layer enables a node to be wise to a period of information from its neighborhood explicitly. We leverage attention mechanisms to select influential neighboring conditions of a node. Instead of attending the central node's current state with its neighbors’ concurrent states, we attend a node’s current state over each of its neighbors’ historical states within a certain period of time. By building connections from each neighbor’s past to the central node’s present,  a node’s neighborhood information no matter in the past or in the present is traversed efficiently and effectively. 

%is capable of capturing the latent dependency between a node's current state and its neighbors' historical states. 

The main contributions of the paper are as follows:
\begin{itemize}
    \item We propose TraverseNet, a simple and powerful framework that captures the inner spatial-temporal dependencies without compromising spatial-temporal integrity.
    %We present that space-time continuum and space-time integrity are vital to learn the spatial-temporal structure data;
    \item We propose \textcolor{black}{a} message traverse layer,  effectively unifying space and time in message passing by traversing information of a node's neighbors' past to the node's present.
    \item We construct TraverseNet with message traverse layers and validate the significance of \textcolor{black}{the} message traverse mechanism with an experimental study.
\end{itemize}

% Drawbacks of previous works 
% \begin{itemize}
% \item Previous works model spatial correlations and temporal correlations separately.  In fact,  a node’s current state may depend on its neighbors' past states within a certain period of time. For example, a traffic congestion of a road will cause another traffic congestion of its nearby nodes in five minutes later. This suggests that treating spatial correlations and temporal correlations separately is not appropriate. 
% \item Many of existing methods assumed fixed edge weights. In some situations, edge weights should be dynamic. Although several works proposed attention-based spatial-temporal networks, they treat spatial correlations and temporal correlations separately and are not able to answer the question at what time a node’s neighbor will affect itself.  
% \item Models of previous works are over-complex.  Various techniques from deep learning are combined together, such as inception, dilated convolution, and ensemble. It is hard to evaluate the real contribution of each component and hard to understand the model in a whole. 

% \end{itemize}

\section{Background and Related Work}
%In this section, we provide background knowledge about spatial-temporal graph modeling and summarize related works with the least amount of technical details.

\subsection{Definitions and Notations}
An attributed graph is defined as $G$ = $(V,E,\mathbf{X})$, where $V$ is the set of nodes, $E$ is the set of edges, and $\mathbf{X}$ is a node feature matrix. We let $v \in V$ to represent a node and $e =(v,u) \in E$ to denote an directed edge from $u$ to $v$. The neighborhood of a node $v$ is the set of nodes $N(v)= \{u\in V| (v,u)\in E\}$ that points to node $v$. The adjacency matrix $\mathbf{A}$ is a mathematical way that defines a graph with a sparse matrix. It is an $N$ by $N$ matrix with $A_{ij} = 1$ if $(v_i,v_j) \in E$ and $A_{ij} = 0$ if $(v_i,v_j) \notin E$, where $N$ is the number of nodes. \textcolor{black}{We define a spatial-temporal graph as a sequence of graphs that evolve over time, $\mathcal{G}_{t_{i-p}:t_i}=\{G_{t_{i-p}}, G_{t_{i-p+1}}, \cdots, G_{t_i}\}$. In this paper, we assume $G_{t_i}=(V,E,\mathbf{X}_{t_i})$, where $\mathbf{X}_{t_i}\in \mathbf{R}^{N\times D}$ denote the node feature matrix at time step $t_i$ and $D$ is the feature dimension.}

%Conditioned on a graph $G=(V,E)$ and a time period from $t_p$ to $t_q$, 
% we define a spatial temporal graph as $G_{st}=(\Upsilon,\Xi; G,t_p,t_q)$. The node set of the spatial temporal graph is $\{v_t \in \Upsilon\}$ where $\{v \in V \}$ and $t \in \{t_p,t_{p+1},\cdots, t_q$. The edge set of the spatial temporal graph is $\{(v_{t},u_{t-o}) \in \Xi \}$ where $t<=t_q$, $t-o>=t_p$, and $(v,u)\in E$.

%We let $\mathbf{X}_t\in \mathbf{R}^{N\times D}$ to denote the node feature matrix at time step $t$, where $N$ is the number of nodes and $D$ is the feature dimension.

\textcolor{black}{The goal of spatial-temporal graph forecasting is to predict future spatial-temporal graphs given its historical data. Formally, the spatial-temporal graph forecasting problem is defined as finding a mapping from the historical values to the future values:
\begin{equation}
    \mathcal{G}_{t_{i-p}:t_i} \xrightarrow[]{f} \mathcal{G}_{t_{i+1}:t_{i+q}}.
\end{equation}
The major notations in this paper are listed in Table \ref{tab:notations}. 
}
% Using our definition of spatial-temporal graph, we could write this problem as 
% \begin{equation}
%     [\mathbf{\chi}_{in};G_{st}]\xrightarrow[]{f}[\mathbf{\chi}_{out}]
% \end{equation}
% where $\mathbf{\chi}_{in} \in \mathbf{R}^{Np\times D} = [\mathbf{X}_{t_0},\mathbf{X}_{t_1},\cdots, \mathbf{X}_{t_p}]$ and $\mathbf{\chi}_{out} \in \mathbf{R}^{Nq\times D} = [\mathbf{X}_{t_{p+1}},\mathbf{X}_{t_{p+2}},\cdots \mathbf{X}_{t_{p+q}}]$.

	\begin{table}[t]
		\caption{\textcolor{black}{Notations.}}
		\label{tab:notations}
		\centering
		\begin{tabular} {  l l p{7cm} } \toprule
				\textbf{Notations}& \textbf{Descriptions} \\ \midrule
				$G$& A graph. \\ \hline
				$V$& The set of nodes in a graph.\\ \hline
				$v$ & A node $v\in V$. \\ \hline
			    $E$& The set of edges in a graph.\\ \hline
				$N(v)$ & The neighbors of a node $v$. \\ \hline
				$\mathbf{A}$ & The graph adjacency matrix.  \\ \hline

				$\mathbf{X}$ & The feature matrix of a graph. \\ \hline
				$\mathbf{X}_t$ & The node feature matrix of a graph at the time step $t$. \\ \hline
				$\mathbf{H}$ & The node hidden feature matrix. \\ \hline
				$\mathbf{h}_v$ & The hidden feature vector of node $v$. \\ \hline

            	$\mathbf{W},\mathbf{\Theta}, \mathbf{U}, \mathbf{\gamma}$ & Learnable model parameters. \\  
				\bottomrule
		\end{tabular}
	\end{table}

\subsection{Graph Neural Networks (GNNs)}
Graph neural networks extract high-level representations of nodes for graphs by graph convolution \cite{kipf2017semi}, \textcolor{black}{\cite{ levie2017cayleynets, bianchi2021graph}}, or message passing \cite{hamilton2017inductive,gilmer2017neural}  in (semi) supervised or self-supervised learning manners \cite{wu2020comprehensive, wan2021contrastive, liu2021graph, zhang2022trustworthy}. Graph convolution methods are based on the eigen-decomposition of the graph Laplacian matrix. The motivation of graph convolution methods is to remove noise from graph signals. Beyond that, many graph convolution methods can be interpreted from the perspective of message passing. Message passing methods aggregate a node's information with its neighborhood information in order to extract its latent representation.

% The general form of a graph convolution/message passing layer is written as:

% \begin{equation}
% \mathbf{h}_v^{(k)} = U_k\Big(\mathbf{h}_v^{(k-1)},\sum_{u\in N(v)} M_k\big(\mathbf{h}_v^{(k-1)},\mathbf{h}_u^{(k-1)}\big)\Big),
% \end{equation}
% where $U_k$ is a update function and $M_k$ represents a message function.  The message function can be a sum or mean function, which assumes the equal contribution of a node's neighbors. 

Monti et al. \cite{monti2017geometric} introduce node pseudo-coordinates to learn the relative weight between a node and its neighbor. Velickovic et al. \cite{velickovic2017graph} propose graph attention to update the contribution weights of a node's neighbors. Gao et al. \cite{gao2019graph} further improve graph attention by only attending to a node's important neighbors. Klicpera et al. \cite{klicpera2019diffusion} design message passing from the perspective of diffusion. It assumes the received information for each node stabilizes after infinite steps of information propagation. Therefore, an approximated diffusion matrix can be utilized in message passing, which broadens a node's receptive field largely. The weights of a diffusion matrix are fixed according to the graph structure.  Wang et al. \cite{wang2020direct} compute the attention scores between pairs of nodes and diffuse node information based on the attention scores until convergence.

\subsection{Spatial-temporal Graph Neural Networks (STGNNs)}
Standard graph neural networks assume that the input node features are static and only consider spatial information flow. When the node features dynamically change over time, a group of methods under the name of \emph{spatial-temporal graph neural networks} can handle the data more effectively \cite{seo2018structured,li2018diffusion,jin2022multivariate}. We divide existing spatial-temporal graph neural networks into three categories: recurrent-based methods, convolution-based methods, and attention-based methods.  
%In the following, we will review methods under each category.

\subsubsection{Recurrent-based STGNNs}
Recurrent-based STGNNs simply assume a node’s current hidden state depends on its own current inputs,  its neighbors’ current inputs, its own previous hidden states and its  neighbors’  previous hidden states \cite{seo2018structured,li2018diffusion,zhang2018gaan}. The form of recurrent-based approaches can be conceptualized as
\begin{equation}
    \mathbf{H}_t = RNN(GCN([\mathbf{X}_t,\mathbf{H}_{t-1}],\mathbf{A};\mathbf{\Theta});\mathbf{U})
\end{equation}
where $\mathbf{\Theta}$ and $\mathbf{U}$ are model parameters, and $\mathbf{H}_{t-1}$ represents the nodes' previous hidden state matrix.  The previous hidden state of a node is essentially a memory vector of its historical information. As the number of recurrent steps increases, the memory vector will gradually forget information many steps before. 
Seo et al. \cite{seo2018structured} are the first to propose a recurrent-based STGNN. They adopt the Long-short Term Memory Networks (LSTMs) as the RNN component and ChebNet as the GCN component. Li et al. \cite{li2018diffusion} consider information diffusion in designing their framework. They replace LSTMs with Gated Recurrent Units (GRUs) and propose diffusion graph convolution. Zhang et al. \cite{zhang2018gaan} further propose multi-head graph attention as the GCN part.
Considering node data distributions may differ, Pan et al. \cite{pan2019urban} introduce a hyper-network that is able to generate a set of STGNN model parameters for each node based on their static node features.
The common drawbacks of recurrent-based STGNNs are the high computation cost and gradient diminishing problem induced by recurrent propagation accompanied \textcolor{black}{by} graph convolution. 

\subsubsection{Convolution-based STGNNs}
Convolution-based STGNNs take advantage of the efficiency and shift-invariance property of convolutional neural networks \cite{li2019spatio,wu2019graph,wu2020connecting, zhang2020spatio}. 
They interleave temporal convolutions with graph convolutions to handle temporal correlations and spatial dependencies respectively. The core difference to recurrent-based methods is that they replace recurrent neural networks with temporal convolution networks (TCN) for capturing temporal patterns, as illustrated \textcolor{black}{in} the following, 
\begin{equation}
    \mathbf{H} = GCN(TCN(\parallel_{t=1}^T \mathbf{X}_t;\mathbf{\Theta}),\mathbf{A};\mathbf{U})
\end{equation}
where $\parallel$ represents concatenation, $\mathbf{H}\in R^{N\times D\times (T-c+1)}$, $T$ is the sequence length, and $c$ is the kernel size of the temporal convolution. The information flow of an STGNN layer occurs first temporally \textcolor{black}{and} then spatially. Such models are against nature, where an object moves in space and time simultaneously.
Li et al. \cite{li2019spatio} firstly propose a CNN-based STGNN. They adopt standard 1D convolution to capture temporal dependencies and Chebnet to capture spatial dependencies. The use of standard 1D convolution is inefficient to handle long-term dependencies. Wu et al. \cite{wu2019graph} propose Graph WaveNet that \textcolor{black}{adopts} dilated 1D convolution. They further propose a self-adaptive graph learning module that can learn latent spatial dependencies from data. While Graph WaveNet only learns static graph structures, Zhang et al. \cite{zhang2020spatio} consider learning both static and dynamic, global and local graph structures.

%\sout{It is against the nature that an object moves in space and time simultaneously} \quan{Such models are against the nature, where an object moves in space and time simultaneously}.

\subsubsection{Attention-based STGNNs}
Similar to convolution-based approaches, attention-based STGNNs treat spatial dependencies and temporal correlations in separate steps \cite{guo2019attention, zheng2020gman, park2020stgrat,wang2020traffic},

\begin{equation}
    \mathbf{H} = SA(\parallel_{t=1}^T TA(\mathbf{X}_t;\mathbf{\Theta}),\mathbf{A});\mathbf{U})
\end{equation}
where $SA(\cdot)$ is a spatial attention layer and $TA(\cdot)$ is an temporal attention layer. 
The motivation of attention-based methods is that node spatial dependency and temporal dependency could be dynamically changing over time. The spatial attention layer first updates the graph adjacency matrix by computing the distance between a query node's input and a key node's input, then \textcolor{black}{performing} message passing. The temporal convolution layer computes a weighted sum of a node's historical state based on attention scores. 
Guo et al. \cite{guo2019attention} propose ASTGCN that interleaves spatial attention with temporal attention operations. To increase the scalability of spatial attention, Zheng et al. \cite{zheng2020gman} propose hierarchical spatial attention that splits nodes into groups and performs inter-group attention and intra-group attention. Later works \cite{park2020stgrat,wang2020traffic} on attention-based STGNN exploit the idea of Transformer \cite{vaswani2017attention} for designing attention-based STGNNs.

\subsubsection{Other relevant works} There are several prior works \textcolor{black}{that} tackle spatial-temporal dependencies jointly. Song et al. \cite{song2020spatial} propose a localized spatial-temporal graph convolution network (STSGCN) that synchronously capture consecutive local spatial-temporal correlations. Li et al. \cite{mengzhang2020spatial} propose the Spatial-Temporal Fusion Graph Neural Networks (STFGNN) that considers pre-defined spatial dependencies, pre-computed sequence similarities, and local temporal dependencies.
Both STSGCN and STFGNN \textcolor{black}{are} not efficient to transfer the historical information of a node's neighbor to the node itself due to local connections and fixed dependency weights. \textcolor{black}{Most recently, Pan et al. \cite{pan2020spatio} and Hadou et al. \cite{hadou2021space} study stability of STGNNs. They design STGNNs that are stable to small perturbations in the underlying graphs. Pan et al. conduct experiments on human action recognition tasks \cite{pan2020spatio}. Hadou et al. validate their method on flocking and motion planning \cite{hadou2021space}. Both of them do not consider traffic forecasting. Isufi et al. propose GTCNN based on graph product \cite{isufi2021graph}. While enabling spatial-temporal dependencies, graph product is limited to fixed edge weights and undirected spatial-temporal relations.}

\textcolor{black}{In a broader context, our paper and previously discussed related work only focus on predicting the trend of node features for spatial-temporal graphs. There are several research that conduct a more general study in which
a full graph (nodes, edges, and attributes) is predicted \cite{zambon2019autoregressive, paassen2020graph}.}

\section{TraverseNet}
To enable the direct flow of information both in space and time, we propose a novel spatial-temporal graph neural network named TraverseNet. The proposed design of TraverseNet is simple. Apart from multi-layer perceptrons (MLPs), TraverseNet only contains our newly proposed message traverse layers. In the following, we introduce the message traverse layer and present the model framework of TraverseNet.

\subsection{Message Traverse Layer}
\begin{figure}
	\centering
	\scalebox{0.5}{\includegraphics[width=\textwidth]{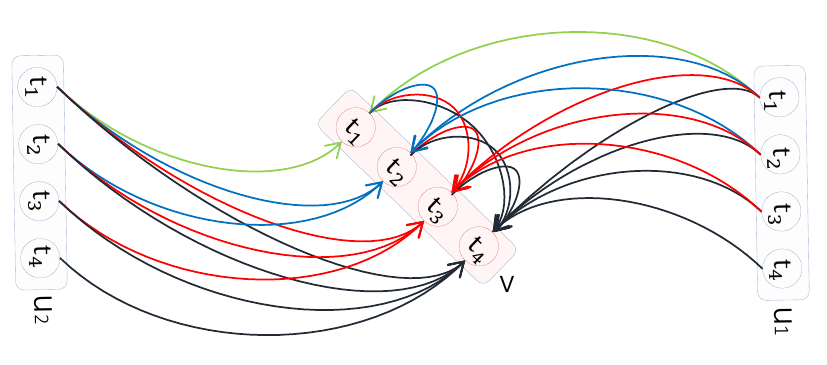}}
	\caption{A demonstration of message traverse layer. The node $v$ has two neighbors $u_1$ and $u_2$. Each node has four consecutive states at $t_1$, $t_2$, $t_3$, $t_4$. The message traverse layer allows the node $v$ at a certain time step to receive information from its own previous time steps as well as its neighbors' previous time steps.} 
	\label{fig:mtlayer}
\end{figure}

\begin{figure*}
	\centering
	\scalebox{0.9}{\includegraphics[width=\textwidth]{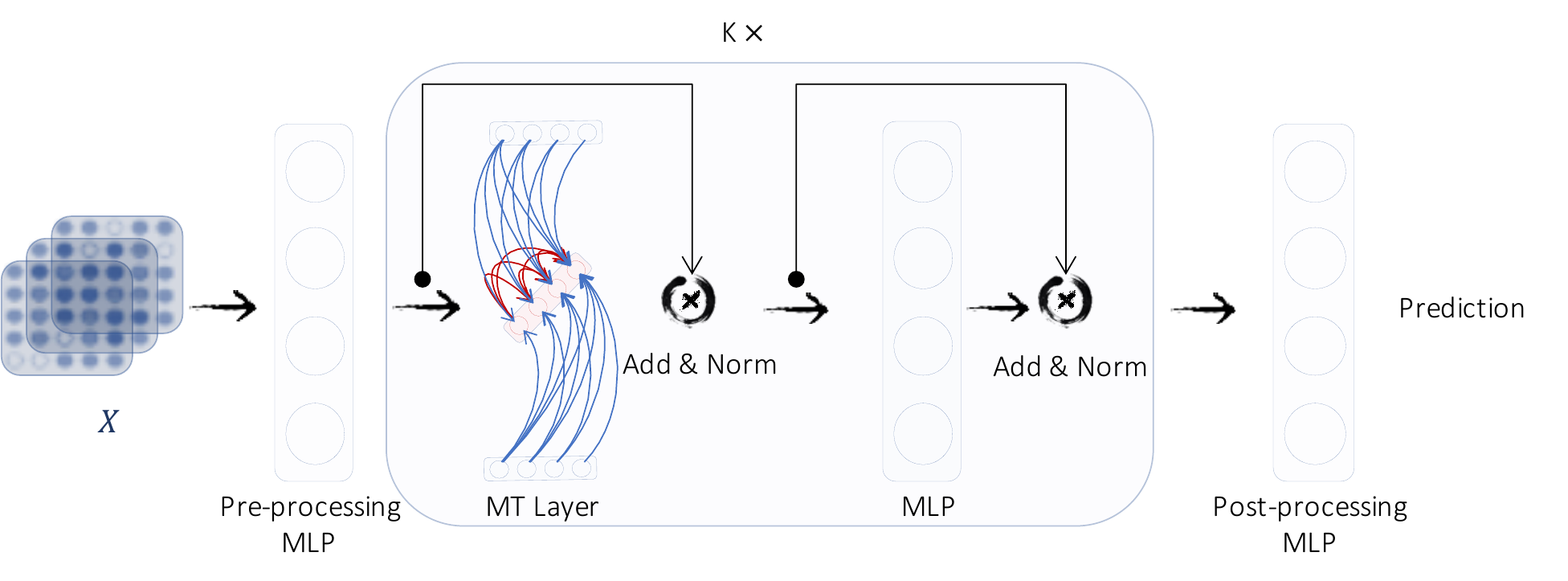}}
	\caption{The model framework of TraverseNet. The TraverseNet mainly consists of three parts, the pre-processing layer, the message traverse layer, and the post-processing layer. The input is a sequence of node feature matrix $\mathbf{X}_{t_1}$, $\mathbf{X}_{t_2}$, $\cdots$, $\mathbf{X}_{t_p}$. The pre-processing MLP layer projects the input feature matrices to a latent feature space. The message traverse layer propagates information across space and time. The post-processing layer projects the nodes' hidden states to the output space. }  
	\label{fig:arch}
\end{figure*}

Static graph convolution layers or message passing layers only pass nodes' neighborhood information across space, ignoring temporal dynamics. We propose the message traverse layer, a message passing layer that allows node neighborhood information to be simultaneously delivered across space and time. As demonstrated \textcolor{black}{in} Figure \ref{fig:mtlayer}, a node $v$ has two neighbors $u_1$ and $u_2$. States of $u_1$ prior to $t_4$ can traverse to the state of node $v$ at $t_4$ directly. This is in fundamental contrast to existing spatial-temporal works, where node $v$ at $t_4$ can only receive information of node $u_1$ and $u_2$ from their concurrent time step $t_4$. The spatial-temporal dependency between two nodes may change from time to time depending on their states. We further leverage the attention mechanism to select important neighborhood information and to handle dynamic spatial-temporal dependencies. Formally, the message traverse layer updates the state of node $v$ for each time step from $t=0$ to $t=p-1$ by

\begin{equation}
    \label{eqn:h}
    \mathbf{h}_{v_t}^{(k)} = \sum_{u\in {N(v)\cup v}}\alpha_r^{(k)}(\mathbf{c}_{(v\to v)_{t}}^{(k)},\mathbf{c}_{(u\to v)_{t}}^{(k)})\mathbf{W}_s^{(k)}\mathbf{c}_{(u\to v)_{t}}^{(k)}, \\
\end{equation}
where $\mathbf{h}_{v_t}^{(0)}=\mathbf{x}_{v_t}$ and $\mathbf{W}_s^{(k)}$ represent learnable model parameters at the $k^{th}$ message traverse layer. The function $\alpha_r^{(k)}(\cdot,\cdot)$ is an attention function of the form
%where for $i \in \{r,c,e\}$,
\textcolor{black}{
\begin{equation}
    \label{eqn:alpha}
    \alpha_r^{(k)}(\mathbf{z}_v,\mathbf{z}_u) = \frac{
exp(\sigma(\mathbf{\gamma}_r^{(k)^T}[\mathbf{\Theta}_{r1}^{(k)}\mathbf{z}_v||\mathbf{\Theta}_{r2}^{(k)}\mathbf{z}_u]))}{\sum_{\mathbf{m}\in S} exp(\sigma(\mathbf{\gamma}_r^{(k)^T}[\mathbf{\Theta}_{r1}^{(k)}\mathbf{z}_v||\mathbf{\Theta}_{r2}^{(k)}\mathbf{z}_m]))},
\end{equation}
}
\textcolor{black}{where $S=\{N(v)\cup v\}$, $\mathbf{\Theta}$, $\mathbf{\gamma}$ represent model parameters, $\mathbf{z}_v$ and $\mathbf{z}_u$ denote the input node hidden features of the attention function, and $\sigma(\cdot)$ denote an activation function.}
The term $\mathbf{c}_{(v\to v)_{t}}^{(k)}$ represents the latent information received by node $v$ from its own time steps prior to time $t$. It is calculated as a weighted sum of its own historical states
\begin{equation}
    \label{eqn:cv}
    \mathbf{c}_{(v \to v)_{t}}^{(k)} = \sum_{m=0}^Q\alpha_{c}^{(k)}(\mathbf{h}_{v_{t}}^{(k-1)},\mathbf{h}_{v_{t-m}}^{(k-1)})\mathbf{W}_c^{(k)}\mathbf{h}_{v_{t-m}}^{(k-1)},  
\end{equation}
The term $\mathbf{c}_{(u\to v)_{t}}^{(k)}$ denotes the latent information received by node $v$ from its neighbor $u$'s previous time steps prior to time $t$. To assess the importance of each state of neighbor $u$, we involve the state of node $v$ at the current time step as a query in the attention function
\begin{equation}
    \label{eqn:cuv}
    \mathbf{c}_{(u\to v)_{t}}^{(k)}  = \sum_{m=0}^Q\alpha_{e}^{(k)}(\mathbf{h}_{v_{t}}^{(k-1)},\mathbf{h}_{u_{t-m}}^{(k-1)})\mathbf{W}_e^{(k)}\mathbf{h}_{u_{t-m}}^{(k-1)}.
\end{equation}
The attention functions $\alpha_c^{(k)}(\cdot,\cdot)$ and $\alpha_e^{(k)}(\cdot,\cdot)$ have the same form as $\alpha_r^{(k)}(\cdot,\cdot)$. The hyper-parameter $Q$ controls the time-window size within which a node receives information from its neighbor's past states. We differentiate the node itself from its neighborhood set because the node's own information has a decisive influence on its predictions for sequence forecasting. The computation complexity of the proposed message traverse layer is $O(M\times p\times Q)$, where $M$ denotes the number of edges including self-loops and $p$ is the input sequence length.

Our message traverse layer is a generalization of both spatial attention layers and temporal attention layers. Specifically, if the neighborhood set of the node $v$ is empty, there will be only one term in the summation in Equation 6 with the attention weight being one.  Equation~\ref{eqn:h} reduces to a \textbf{temporal attention layer}
\begin{equation}
\mathbf{h}_{v_{t}}^{(k)} = \mathbf{W}_s^{(k)} \sum_{m=0}^Q\alpha_{c}^{(k)}(\mathbf{h}_{v_{t}}^{(k-1)},\mathbf{h}_{v_{t-m}}^{(k-1)})\mathbf{W}_c^{(k)}\mathbf{h}_{v_{t-m}}^{(k-1)}.
\end{equation}
Similarly, if the window size $Q$ is set to 0, then Equation \ref{eqn:h} becomes a \textbf{spatial attention layer}
\begin{equation}
    \mathbf{h}_{ v_t}^{(k)} = \sum_{u\in {N(v)\cup v}}\alpha_r^{(k)}(\mathbf{W}_c^{(k)}\mathbf{h}_{v_{t}}^{(k-1)} ,\mathbf{W}_e^{(k)}\mathbf{h}_{u_{t}}^{(k-1)})\mathbf{W}_s^{(k)}\mathbf{c}_{(u\to v)_{t}}^{(k)},
\end{equation}
In contrary to existing works that interleave spatial computations with temporal computations, the proposed message traverse layer handles spatial-temporal dependency as a whole. 
Our message traverse layer can not be separated to a spatial attention layer and a temporal attention layer. It is mainly because we assume the spatial-temporal dependency between a node's state at time step $t$ and its neighbor's state at time step $t-m$ is dynamic. 
We use the state of the central node $v$ at time step $t$ as a query to assess the importance of its neighboring node $u$'s historical states at each time step. Overall this design shortens the path length of message passing and enables a node to be aware of its neighborhood variation at firsthand.

\subsection{Model Framework}
As the message traverse layer is sufficient to capture spatial-temporal dependencies, we design a framework named TraverseNet that is simple and powerful to accomplish the spatial-temporal graph forecasting task. 
In Figure \ref{fig:arch}, we present the framework of TraverseNet.
The TraverseNet consists of three parts, the pre-processing layer, a stack of message traverse layers, and the post-processing layer. The pre-processing layer is a feedforward layer \textcolor{black}{that} projects node feature matrix at each time step to a latent space. Next, we capture nodes' spatial-temporal dependencies by message traverse layers. Finally, we 
use the post-processing layer to map node hidden states to the output space. The post-processing layer contains a $1\times p$ standard convolutional layer followed by a feedforward layer. Suppose the input of node $v$ to the post-processing layer is $\mathbf{z}_v= \parallel_{i=0}^{p-1} \mathbf{h'}_{v_i}^{(K-1)}$, where $\mathbf{z}_v \in \mathbf{R}^{d\times 1 \times p}$, and $p$ is the input sequence length. The $1\times p$ standard convolutional layer is used to squeeze the third dimension of the input $\mathbf{z}_v$ to 1. Afterward, the feedforward layer is applied to generate the prediction $\mathbf{z}_v'\in \mathbf{R}^q$ for node $v$, where $q$ is the output sequence length. In addition, residual connections and batch normalization are applied to message traverse  layers to improve model robustness. In particular, as the input of each node may have very different scales, we let the batch normalization scale the hidden features on the node dimension. 

\subsection{Optimization \& Implementation}
We optimize model parameters of TraverseNet end-to-end by minimizing the Mean Absolute Error (MAE) loss with gradient descent. The MAE is defined as 
\begin{equation}
    L = \mathrm{Average}\left(\sum_{i=p+1}^{p+q}|\mathbf{X}_{t_i}-\hat{\mathbf{X}}_{t_i}|\right).
\end{equation}
We implement TraverseNet with Pytorch and DGL \cite{wang2019dgl}. In more detail, we first construct a heterogeneous graph by treating each node at each time step as a unique node and creating connections as illustrated \textcolor{black}{in} Figure \ref{fig:mtlayer}.  The state of each node at a certain time step is linked to its historical states as well its neighbors' historical states within a time window $Q$. 
The constructed graph is in a sparse form thus efficient for computation. We implement the message traverse layer by customizing the HeterGraphConv module of DGL. The code is publicly available at \href{https://github.com/nnzhan/TraverseNet}{https://github.com/nnzhan/TraverseNet}.

\section{Experimental Studies}

\subsection{Dataset}

\begin{table}
\begin{threeparttable}
\small
\caption{Dataset statistics. \label{tab:stat}}
\begin{tabular}{lllll}
\toprule
Datasets & \# Sensors & Sampling rate & \# Time steps & Signals\\
\midrule
PEMS03   & 358               & 5 mins          & 26209   & F             \\
PEMS04   & 307               & 5 mins         & 16992   & F,S,O             \\
PEMS08   & 170               & 5 mins          & 17856  & F,S,O\\
\bottomrule
\end{tabular}
\footnotesize
In column titled ``Signals'', F represents traffic flow, S represents traffic speed, and O represents traffic occupancy rate.
\end{threeparttable}
\end{table}

\begin{figure*}[ht]
        % \begin{subfigure}[b]{0.3\textwidth}   
        %     \centering 
        %     \includegraphics[width=\textwidth]{./FIG/explot/cor03.pdf}
        %     \caption[PEMS-03]%
        %     {{\small PEMS-03}}    
        %     \label{fig:cor3}
        % \end{subfigure}
        % \hfill
        \begin{subfigure}[b]{0.5\textwidth}   
            \centering 
            \includegraphics[width=\textwidth]{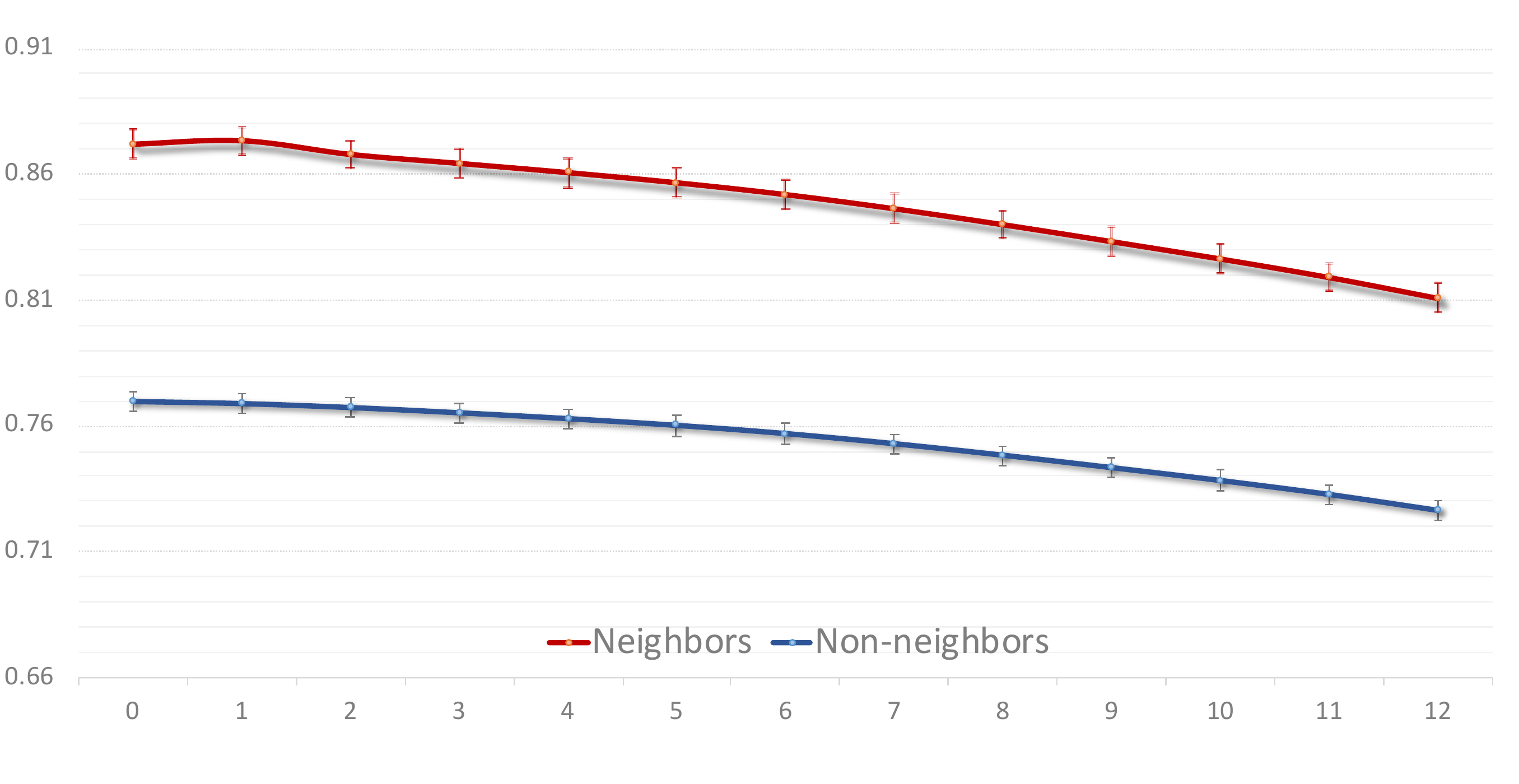}
            \caption[PEMS-04]%
            {{\small PEMS-04}}    
            \label{fig:cor4}
        \end{subfigure}
        \hfill
        \begin{subfigure}[b]{0.5\textwidth}   
            \centering 
            \includegraphics[width=\textwidth]{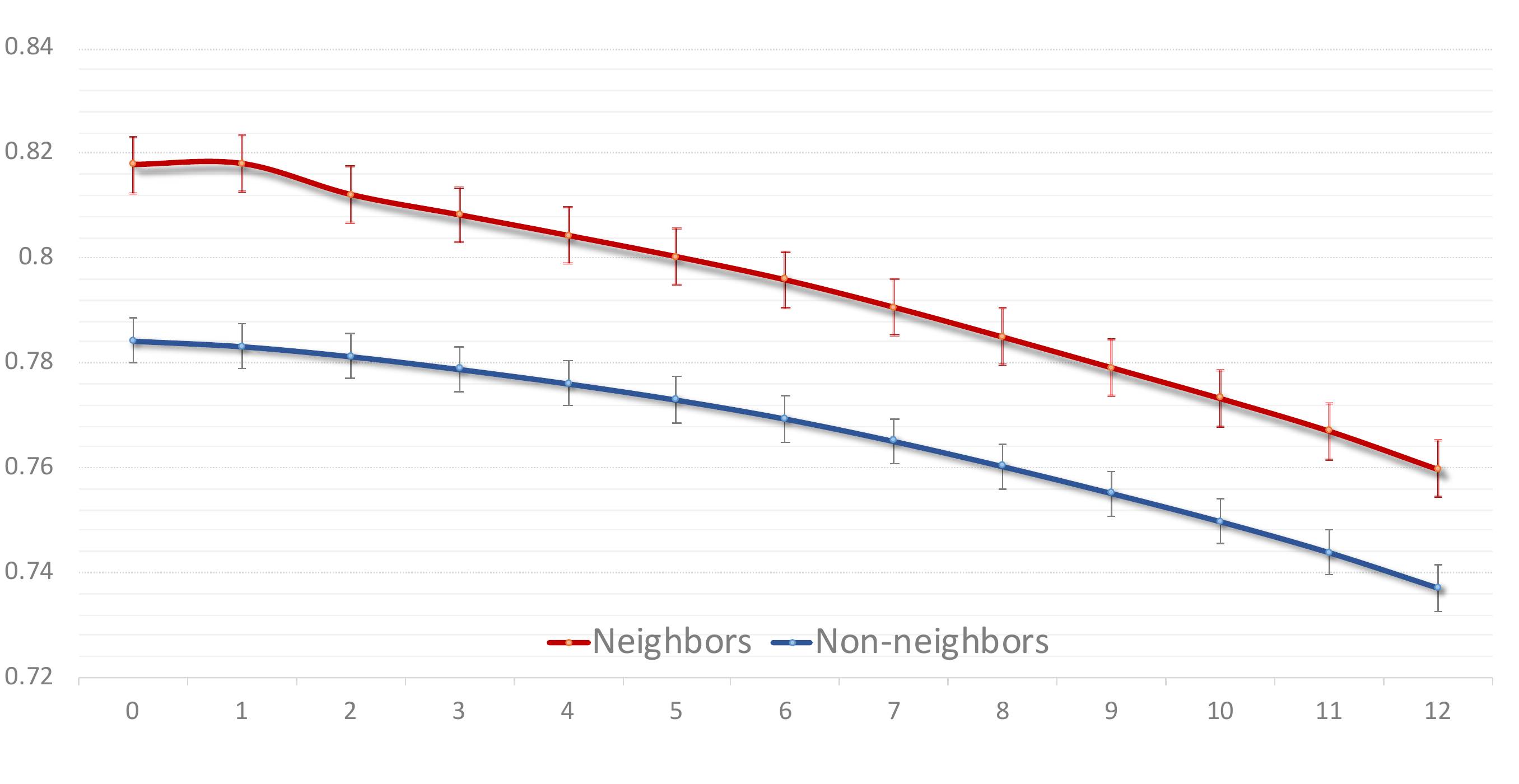}
            \caption[PEMS-08]%
            {{\small PEMS-08}}    
            \label{fig:cor8}
        \end{subfigure}
        \caption[Cross-correlations]
        {\small Cross-correlations between pairs of connected nodes and between pairs of far-away nodes. The x-axis is the time lag. The y-axis is the mean of correlation coefficients with standard deviation. Red lines denote cross-correlations between pairs of connected nodes. Blue lines denote cross-correlations between pairs of far-away nodes. } 
        \label{fig:cross}
\end{figure*}
\begin{figure*}
        %         \begin{subfigure}[b]{0.3\textwidth}   
        %     \centering 
        %     \includegraphics[width=\textwidth]{./FIG/explot/dist03.pdf}
        %     \caption[PEMS-03]%
        %     {{\small PEMS-03}}    
        %     \label{fig:dis3}
        % \end{subfigure}
        % \hfill
        \begin{subfigure}[b]{0.5\textwidth}   
            \centering 
            \includegraphics[width=\textwidth]{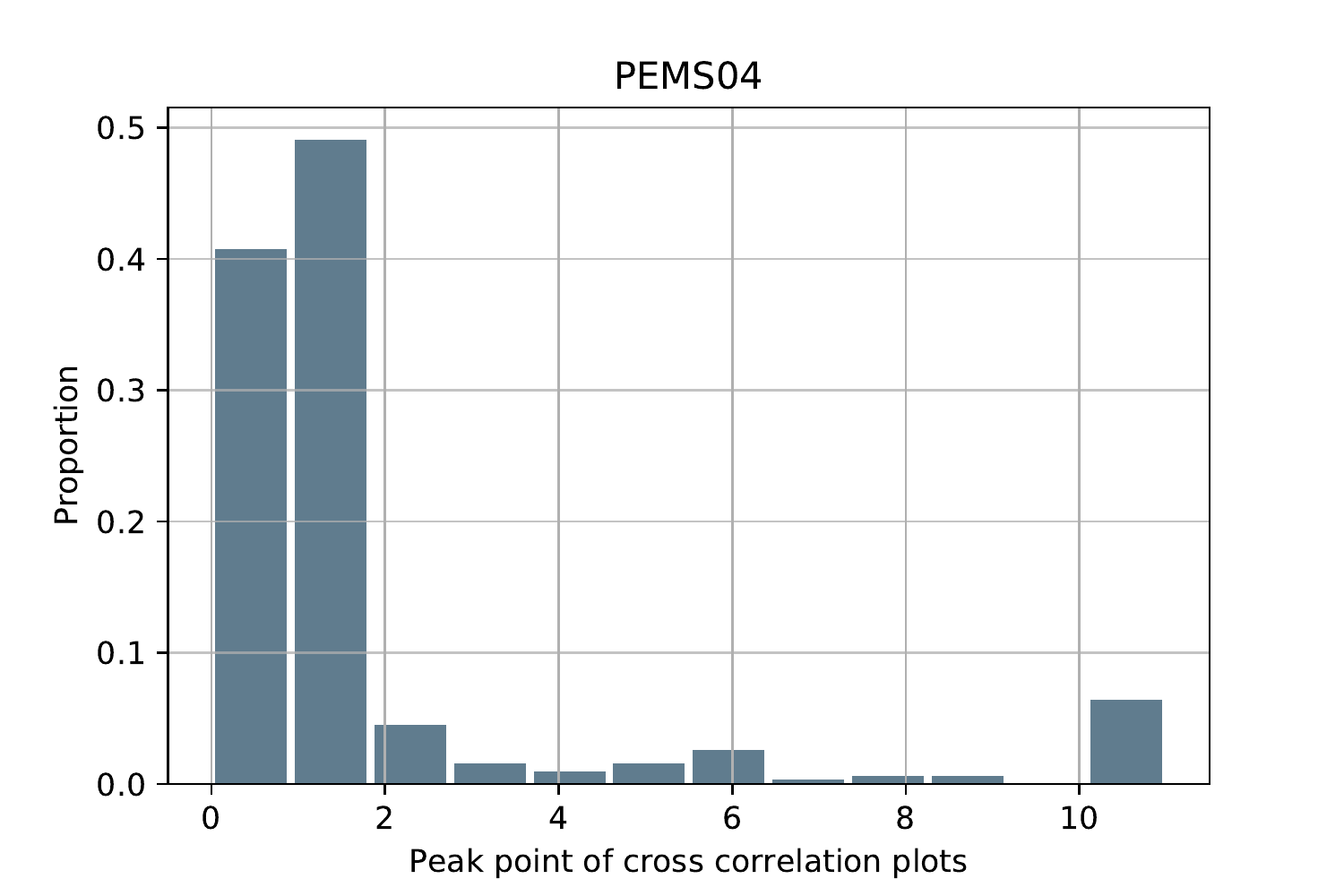}
            \caption[PEMS-04]%
            {{\small PEMS-04}}    
            \label{fig:dis4}
        \end{subfigure}
        \hfill
        \begin{subfigure}[b]{0.5\textwidth}   
            \centering 
            \includegraphics[width=\textwidth]{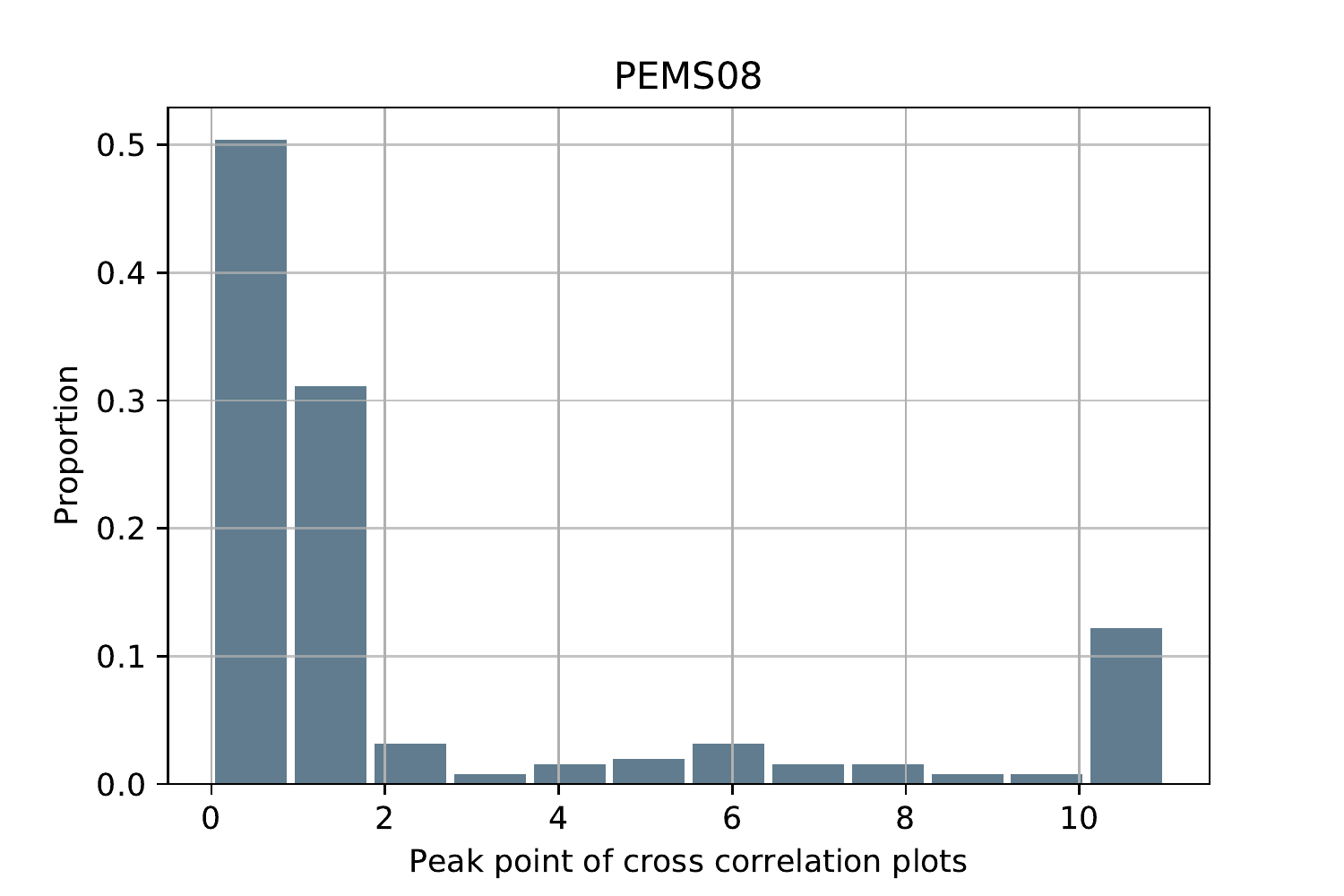}
            \caption[PEMS-08]%
            {{\small PEMS-08}}    
            \label{fig:dis8}
        \end{subfigure}
        \caption[Peak point]
        {\small The distribution of peak points of cross-correlation plots between pairs of connected nodes. The x-axis is the time lag. The y-axis is the proportion. } 
        \label{fig:peak}
\end{figure*}

We follow the experimental setup in \cite{song2020spatial}. We use three traffic datasets, PEMS-03, PEMS-04, and PEMS-08, in our experiments.  These datasets contain traffic signals of road sensors aggregated every five minutes collected by the Caltrans Performance Measurement Systems in different districts of California. We provide summary statistics of each dataset in Table \ref{tab:stat}. Two tasks, i.e. traffic flow prediction and traffic speed prediction, are evaluated using these datasets.
We predict the next twelve steps \textcolor{black}{of} traffic speed/flow given the previous twelve steps of traffic signals and the traffic graph. \textcolor{black}{Note that in traffic forecasting, predicting a sequence length of 12 is a long-term prediction as 12 steps already represent a one-hour period.  The variation of traffic conditions is very large beyond one hour.}  We construct the traffic graph by regarding each sensor as a node and connecting two sensors if they are on the same road.
%quan{I remember that you were using nearest neighbor of travel distance.  This is no longer the case?  Or maybe I forgot the changes.  Also if we have multiple sensors on the same road, you will construct a clique for those nodes?}. 
\textcolor{black}{For} data pre-processing, we standardize the input to have zero mean and unit variance by
\begin{equation}
    \mathbf{\Tilde{X}} = \frac{X-\mathrm{mean}(X)}{\mathrm{std}(X)}.
\end{equation}

To check if the datasets exhibit spatial-temporal dependencies, we plot time lags v.s. cross-correlations of pairs of connected nodes and pairs of far-away nodes (i.e. more than nine hops away) for each dataset respectively. 
%Here we mean pairs of connected nodes as positive entries in a graph adjacency matrix and pairs of non-connected nodes as zero entries in the ninth-order of the graph adjacency matrix.  

%\quan{"Positive entries" and "zero entries" are brand new terms and never used elsewhere.  "Positive entries" are identical to "connected nodes" and "zero entries" are referring to "non-connected nodes", I assume?  If that's the case, you can simply replace "non-connected nodes" with "far-away nodes" and give a definition of "far-away nodes" (ninth-order adjacency matrix entry blah blah blah) and remove this sentence altogether.} 
The cross-correlation between a sequence $\{x_1,x_2,\cdots,x_L\}$ and a sequence $\{y_1,y_2,\cdots,y_L\}$ at time lag $k$ is essentially the correlation between the sequence $\mathbf{y}$ and the sequence $\mathbf{x}$ shifted $k$ steps back:
\begin{equation}
\resizebox{.9\hsize}{!}{$
    C = \frac{\frac{1}{L}\sum_{t=k}^{L}x_{t-k}y_t-\frac{1}{L^2}\sum_{t=k}^{L}x_{t-k}\sum_{t=k}^{L}y_t}{\sqrt{\frac{1}{L}\sum_{t=k}^{L}x_{t-k}^2-(\frac{1}{L}\sum_{t=k}^{L}x_{t-k})^2}\sqrt{\frac{1}{L}\sum_{t=k}^{L}y_{t}^2-(\frac{1}{L}\sum_{t=k}^{L}y_{t})^2}}$}
\end{equation}
Figure \ref{fig:cross} and Figure \ref{fig:peak} \textcolor{black}{present} our analysis. In Figure \ref{fig:cor4} and \ref{fig:cor8}, it shows that the cross-correlations between pairs of connected nodes are always higher than the cross-correlations between pairs of far-away nodes across all time lags and datasets. Digging into detail, we plot the distribution of peak points of cross-correlation curves between pairs of connected nodes for each datasets, as shown \textcolor{black}{in} Figure \ref{fig:dis4}, \ref{fig:dis8}. The majority of two connected nodes’ cross-correlations peak at time lag 0 and 1. However, there is still a small amount of nodes \textcolor{black}{in} which the cross-correlation values peak at higher time lags. In particular, for PEMS-04 and PEMS-08, there are 5\% and 10\% of connected nodes of which the cross-correlation peak at time 11. This suggests that it is reasonable to consider spatial-temporal dependencies explicitly. 

%indicating that it is reasonable to consider spatial-temporal dependencies.

\begin{table*}[tb]
		\centering
%Experiments are repeated for two times and we report the mean and standard deviation.}
		%\vskip -0.1in
		\caption{Performance comparison. \label{tab:result_traffic_speed}}
		\resizebox{\textwidth}{!}{
\begin{threeparttable}
		\begin{tabular}{l l  r r r | r r r |r r r}
			\toprule
			\multirow{2}{*}{Task} & \multirow{2}{*}{Models}   & 
		    \multicolumn{3}{c}{PEMS-03} & \multicolumn{3}{c}{PEMS-04}  & \multicolumn{3}{c}{PEMS-08} \\
			\cline{3-5}  \cline{6-8} \cline{9-11}   
			&& {\small MAE} & {\small MAPE} & {\small RMSE} & {\small MAE} & {\small MAPE} & {\small RMSE} & {\small MAE} & {\small MAPE} & {\small RMSE}\\
			\midrule
			\multirow{8}{*}{\rotatebox[origin=c]{90}{Traffic Flow}}
			& GRU & 20.01 $\pm$ 0.02 & 19.82 $\pm$ 0.06  & 32.52 $\pm$ 0.10 & 24.84 $\pm$ 0.04 & 16.98 $\pm$ 0.09 & 38.87 $\pm$ 0.08 &  18.86 $\pm$ 0.05 & 12.07 $\pm$ 0.05 & 30.25 $\pm$ 0.05\\
            
            & DCRNN & 16.37 $\pm$ 0.05 & \textBF{15.96 $\pm$ 0.10} & 28.37 $\pm$ 0.33 & 24.85 $\pm$ 0.12 & 16.94 $\pm$ 0.16 & 38.95 $\pm$ 0.14 & 17.82 $\pm$ 0.13 & 11.39 $\pm$ 0.10 & 27.69 $\pm$ 0.16  \\
            
            & STGCN & 19.58 $\pm$ 0.15 & 20.17 $\pm$ 0.36 & 32.57 $\pm$ 0.85 & 23.96 $\pm$ 0.07 & 17.40 $\pm$ 0.58 & 36.94 $\pm$ 0.14 & 18.75 $\pm$ 0.15 & 13.00 $\pm$ 0.44 & 28.49 $\pm$ 0.10 \\
            
            & ASTGCN & 18.19 $\pm$ 0.22 & 18.47 $\pm$ 0.54 & 30.58 $\pm$ 0.48 & 22.91 $\pm$ 0.44 & 16.96 $\pm$ 0.54 & 35.60 $\pm$ 0.75 & 18.74 $\pm$ 0.41 & 12.23 $\pm$ 0.30 & 28.80 $\pm$ 0.72 \\
            
            & Graph WaveNet & 16.74 $\pm$ 0.05 & 18.56 $\pm$ 1.66 & 27.75 $\pm$ 0.13 & 20.95 $\pm$ 0.09 & 14.55 $\pm$ 0.17 & 32.64 $\pm$ 0.11 & \textBF{15.66 $\pm$ 0.08} & \textBF{10.31 $\pm$ 0.11} & \textBF{24.59 $\pm$ 0.12} \\
            %& ST-MetaNet & 18.58 $\pm$ 0.10 & 17.84 $\pm$ 0.11 & 31.09 $\pm$ 0.20 & 24.07 $\pm$ 0.11 & 16.55 $\pm$ 0.09 & 37.11 $\pm$ 0.19 & 19.01 $\pm$ 0.19 & 11.90 $\pm$ 0.11 & 29.69 $\pm$ 0.32\\
            & STSGCN & 17.77 $\pm$ 0.20 & 17.28 $\pm$ 0.06 & 28.93 $\pm$ 0.34 & 22.61 $\pm$ 0.07 & 14.90 $\pm$ 0.05 & 35.15 $\pm$ 0.13 & 17.92 $\pm$ 0.14 & 11.60 $\pm$ 0.14 & 27.48 $\pm$ 0.21 \\
            %& 18.01 $\pm$ 0.09 & 17.46 $\pm$ 0.09  & 29.02 $\pm$ 0.59 & 22.68 $\pm$ 0.06 & 14.93 $\pm$ 0.02 & 35.28 $\pm$ 0.10 & 18.21 $\pm$ 0.16 & 11.79 $\pm$ 0.13 & 27.83 $\pm$ 0.21 \\
            & STFGNN & 16.56 $\pm$ 0.32 & \underline{16.09 $\pm$ 0.16} & 28.60 $\pm$ 0.23 & 21.47 $\pm$ 0.10 & \textBF{14.10 $\pm$ 0.08} & 33.57 $\pm$ 0.11 & 17.75 $\pm$ 0.15 & 11.23 $\pm$ 0.09 & 27.64 $\pm$ 0.23 \\
            
            & \textbf{TraverseNet} & \textBF{15.44 $\pm$ 0.10} & 16.41 $\pm$ 0.87 & \textBF{24.75 $\pm$ 0.32} & \textBF{19.86 $\pm$ 0.11} & \underline{14.38 $\pm$ 0.79} & \textBF{31.54 $\pm$ 0.28} & \underline{15.68 $\pm$ 0.12} & \underline{10.87 $\pm$ 0.05} &  \underline{24.62 $\pm$ 0.13} \\
			\midrule
			\multirow{8}{*}{\rotatebox[origin=c]{90}{Traffic Speed}}
			& GRU & - & - & - & 2.34 $\pm$ 0.07 & 5.03 $\pm$ 0.08 & 5.09 $\pm$ 0.02 & 1.80 $\pm$ 0.03 & 3.59 $\pm$ 0.06 & 3.99 $\pm$ 0.04\\
            
            & DCRNN & - & - & - & 2.24 $\pm$ 0.06   & 5.05 $\pm$ 0.33   & 4.89 $\pm$ 0.18  & 1.72 $\pm$ 0.04 & 3.75 $\pm$ 0.15 & 3.75 $\pm$ 0.09 \\
            
            & STGCN & - & - & - & 1.80 $\pm$ 0.01 & 3.87 $\pm$ 0.03 & 4.09 $\pm$ 0.04 & 1.52 $\pm$ 0.01 & 3.28 $\pm$ 0.05 & 3.65 $\pm$ 0.05 \\
            
            & ASTGCN & - & - & - & 1.78 $\pm$ 0.02 & 3.87 $\pm$ 0.12 & 3.97 $\pm$ 0.12 & 1.51 $\pm$ 0.04 & 3.41 $\pm$ 0.10 & 3.67 $\pm$ 0.12 \\
            
            & Graph WaveNet & - & - & - & 1.61 $\pm$ 0.00 & 3.39 $\pm$ 0.02 & 3.71 $\pm$ 0.01 & \textBF{1.34 $\pm$ 0.00} & \textBF{2.97 $\pm$ 0.05} & \textBF{3.36 $\pm$ 0.04} \\
            %& ST-MetaNet & \\
            & STSGCN &  - & - & - & 1.96 $\pm$ 0.06 & 4.28 $\pm$ 0.16 & 4.30 $\pm$ 0.09 & 1.73 $\pm$ 0.07 & 3.80 $\pm$ 0.21 &  3.87 $\pm$ 0.15 \\
            & STFGNN & - & - & - &  1.79 $\pm$ 0.03 & 3.89 $\pm$ 0.08 & 4.03 $\pm$ 0.05 & 1.54 $\pm$ 0.01 & 3.36 $\pm$ 0.02 & 3.62 $\pm$ 0.02 \\
            & \textbf{TraverseNet} & - & - & - & \textBF{1.59 $\pm$ 0.00} & \textBF{3.37 $\pm$ 0.02} & \textBF{3.67 $\pm$ 0.01} & \underline{1.35 $\pm$ 0.01} & \underline{3.02 $\pm$ 0.05} & \underline{3.44 $\pm$ 0.04} \\
			\bottomrule
		\end{tabular}
		Results of best-performing method \textcolor{black}{are} shown in bold font. Results of second-best-performing method \textcolor{black}{are} shown with underlines.
\end{threeparttable}
}
	\end{table*}

\begin{table*}
\small
\caption{Ablation study.}
\label{tab:ablation}
\resizebox{\textwidth}{!}{

\begin{threeparttable}
\begin{tabular}{lcccccc}

\toprule
PEMS-08 &w/o spatial info & w/o st traversing  & w/o attention  & w/o residual & w/o norm & default \\
\midrule
MAE           &  15.95 $\pm$ 0.12  & 15.84 $\pm$ 0.07  &  15.69 $\pm$ 0.11 & 17.01 $\pm$ 0.02   & 15.76 $\pm$ 0.10 & 15.68 $\pm$ 0.12\\
MAPE           & 10.83 $\pm$ 0.20    & 11.05 $\pm$ 0.27   & 10.56 $\pm$ 0.13 & 13.31 $\pm$ 0.90 & 10.86 $\pm$ 0.19 & 10.87 $\pm$ 0.05  \\
RMSE           & 24.99 $\pm$ 0.12    & 24.80 $\pm$ 0.07   & 24.66 $\pm$ 0.15  & 26.24 $\pm$ 0.26 & 24.82 $\pm$ 0.10 & 24.62 $\pm$ 0.13  \\
\toprule
% PEMS-04 &w/o spatial info & w/o st traversing  & w/o attention  & w/o residual & w/o norm & default \\
% \midrule \\
% MAE & 19.93 $\pm$ 0.09 & 20.10 $\pm$ 0.11 & 19.91 $\pm$ 0.12 & 20.62 $\pm$ 0.11 & 20.39 $\pm$ 0.17 & 19.86 $\pm$ 0.11\\
% MAPE & 14.52 $\pm$ 0.23 & 14.63 $\pm$ 0.46 & 14.12 $\pm$ 0.18 & 14.68 $\pm$ 0.35 & 14.16 $\pm$ 0.17 &  14.38 $\pm$ 0.79\\
% RMSE & 31.68 $\pm$ 0.11 & 31.84 $\pm$ 0.22 & 31.86 $\pm$ 0.40 & 32.46 $\pm$ 0.26 & 32.21 $\pm$ 0.29 &  31.54 $\pm$ 0.28\\
% \bottomrule
\end{tabular}
\footnotesize
The results were obtained on the PEMS-08 dataset.
\end{threeparttable}
}
\end{table*}
\subsection{Baseline Methods}
Seven baseline methods are selected in our experiments. Except \textcolor{black}{for} STSGCN and STFGNN, we implement all baseline methods in a unified framework. As it is difficult to merge STSGCN and STFGNN into our framework, we directly use the codes of STSGCN and STFGNN in experiments. We give a short description of each baseline method in the following:

\begin{itemize}
    \item GRU a sequence-to-sequence model \cite{sutskever2014sequence} consists of GRU units \cite{cho2014learning}, not considering spatial dependency.
    %\quan{Is this the way you train it?}.
    \item DCRNN \cite{li2018diffusion} that adopts LSTMs and diffusion graph convolution in a sequence-to-sequence framework.
    \item STGCN \cite{yu2018spatio} that combines gated temporal convolution with graph convolution to capture spatial dependencies and temporal dependencies respectively.
    \item ASTGCN \cite{guo2019attention} that interleaves spatial attention with temporal attention to capture dynamic spatial dependencies and temporal dependencies.
% \end{itemize}
% \begin{itemize}
    \item Graph WaveNet \cite{wu2019graph} that integrates WaveNet with graph convolution. %The self-adpative adjacency matrix is excluded for fair comparison.
    %\item ST-MetaNet \citep{pan2019urban}, a hypernetwork that condition STGNN model parameters \quan{Which parameters?  Temporal or spatial convolution?} on node inputs.
    \item STSGCN \cite{song2020spatial} that considers spatial-temporal dependencies in local adjacent time steps.
    \item STFGNN \cite{mengzhang2020spatial} that considers pre-defined spatial dependencies, pre-computed time series similarities, and local temporal dependencies.
\end{itemize}

\subsection{Experimental Setting}
\label{sec:exptsetting}
We conduct the experiments on the AWS cloud with the p3.8xlarge instance. We train the proposed TraverseNet with the Adam optimizer on a single 16GB Tesla V100 GPU. We split the datasets into train, validation, and test data with a ratio of 6:2:2. We set the number of training epochs to 50, the learning rate to 0.001, the weight decay rate to 0.00001, and the dropout rate to 0.1. We set the number of layers to 3, the hidden feature dimension to 64, and the window size $Q$ to 12. For other baseline methods, we use the default parameters settings reported in their papers.  Each experiment is repeated 5 times  and the mean of evaluation metrics including Mean Absolute Error (MAE), Root Mean Squared Error (RMSE), \textcolor{black}{and} Mean Absolute Percentage Error (MAPE) on test data are reported based on the best model on the validation data.
%\quan{How do you split the train-validation-test set?  And how do you pick the model for evaluating on test set?}.  

\subsection{Overall Results}

Table \ref{tab:result_traffic_speed} presents the main experimental results of our TraverseNet compared with baseline methods. There are missing values on PEMS-03 for the traffic speed prediction because PEMS-03 \textcolor{black}{does} not contain traffic speed information. Among all methods, TraverseNet achieves the lowest MAE and RMSE on PEMS-03 and PEMS-04 for traffic flow prediction and on PEMS-04 for traffic speed prediction. It achieves the second-lowest MAE, MAPE, and RMSE on PEMS-08 for both traffic flow prediction and traffic speed prediction--though the performance gap between the top two methods is extremely small. 

We believe that the reason that Graph WaveNet performs slightly better than TraverseNet on PEMS-08 is that the spatial signal on that dataset is weak. This is supported by the results of the ablation study (cf. Table~\ref{tab:ablation}), which shows that the performance of TraverseNet decreases only slightly when the spatial component of the model is removed. For temporal patterns, the WaveNet component in Graph WaveNet is a very powerful feature extractor for time series data and this is why its performance is somewhat better than TraverseNet. \textcolor{black}{More importantly, TraverseNet significantly outperforms STSGCN and STFGNN. These two methods have the same motivation as our paper which is to jointly handle spatial-temporal dependencies. TraverseNet differs from STSGCN and STFGNN in that they only consider local adjacent spatial-temporal dependencies like RNN-based methods (as demonstrated by Fig \ref{fig:rnn}) while our method could traverse long-range historical information of a node to another node directly (as demonstrated by Fig \ref{fig:our}).}

%To futher investigate the advantage of our model, we split the test data into three groups based on the range of target values. The results are shown in Table \ref{tab:ad}

\subsection{Ablation Study}

We perform an ablation study to validate the effectiveness of the message traverse layer in our model. We are mainly concerned about whether the spatial-temporal message traverse layer is effective and whether the attention mechanism in the message traverse layer is useful. To answer these questions, we compare our TraverseNet model with five different settings listed below:
\begin{itemize}
    \item \textbf{w/o spatial information.} We only use temporal information, which means the neighborhood set of a node is empty.
    \item \textbf{w/o st traversing (without spatial-temporal traversing).} We handle spatial dependencies and temporal dependencies separately by interleaving spatial attention with temporal attention.
    \item \textbf{w/o attention.} We replace attention scores produced by the attention functions with identical weights.
    \item \textbf{w/o residual.} We cancel residual connections for message traversing layers and MLP layers.
    \item \textbf{w/o norm.} We remove batch normalization after message traversing layers and MLP layers. 
\end{itemize} 
We repeat each experiment 5 times. Table~\ref{tab:ablation} reports the mean MAE, MAPE, and RMSE with standard deviation on PEMS-08 test data. We observe that the involvement of spatial information incrementally contributes to model performance. More importantly, spatial-temporal traversing is superior to \textcolor{black}{processing} spatial dependencies and temporal dependencies separately based on the fact the performance of \textbf{w/o st traversing} is lower than the performance of \textbf{default}. \textbf{W/o attention} nearly does not improve model performance, suggesting that the effect of attention mechanisms is limited in time series forecasting.  Besides, according to Table \ref{tab:ablation}, the effectiveness of residual connections and batch normalization is verified.

\begin{figure*}
        \begin{subfigure}[b]{0.3\textwidth}   
            \centering 
            \includegraphics[width=\textwidth]{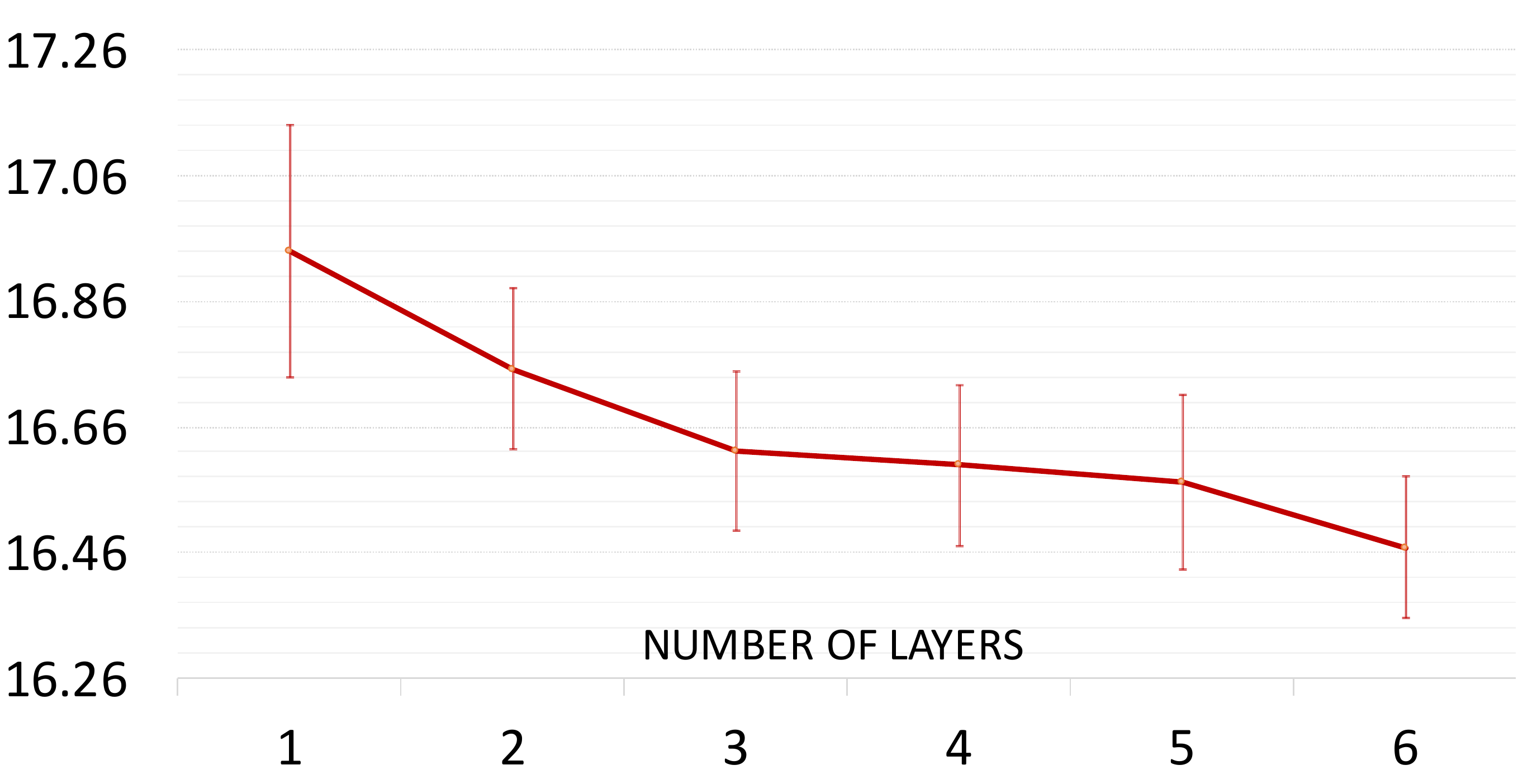}
            \caption[Number of layers]%
            {{\small Number of layers}}    
            \label{fig:7a}
        \end{subfigure}
        \hfill
       \begin{subfigure}[b]{0.3\textwidth}   
            \centering 
            \includegraphics[width=\textwidth]{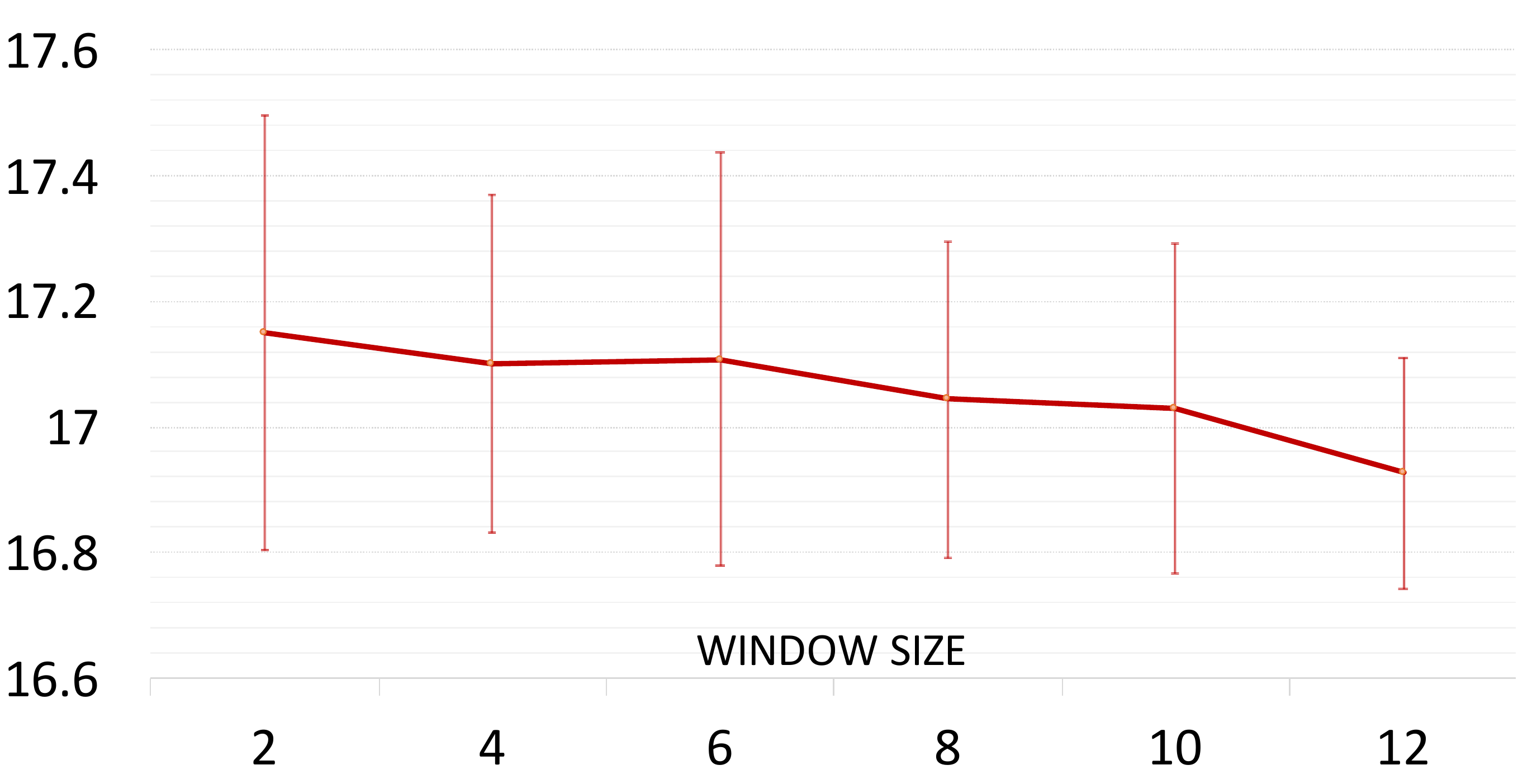}
            \caption[Window size]%
            {{\small Window size}}    
            \label{fig:7c}
        \end{subfigure}
        \hfill
       \begin{subfigure}[b]{0.3\textwidth}   
            \centering 
            \includegraphics[width=\textwidth]{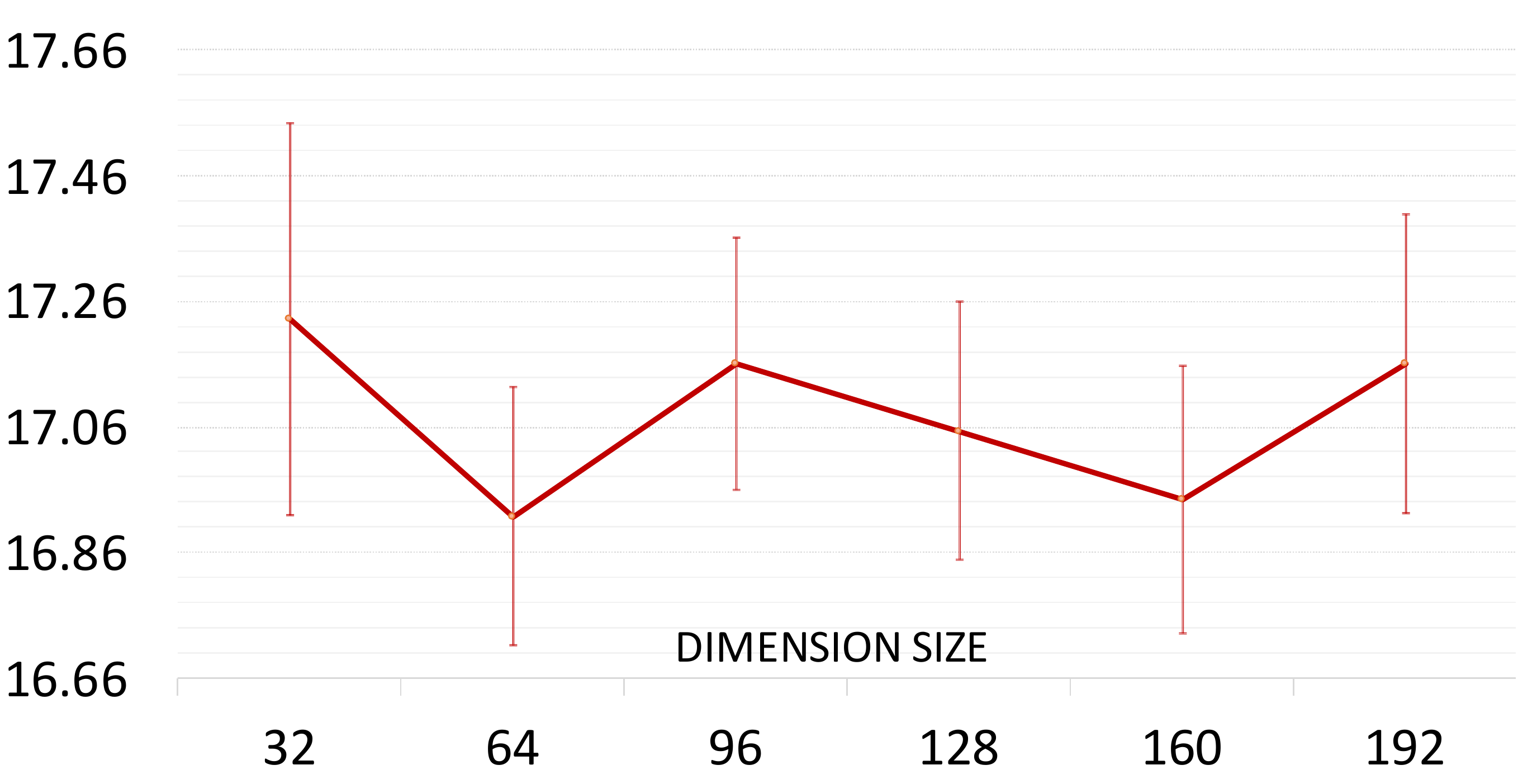}
            \caption[Hidden dimension]%
            {{\small Hidden dimension}}    
            \label{fig:7 b}
        \end{subfigure}
        \newline
        \begin{subfigure}[b]{0.3\textwidth}   
            \centering 
            \includegraphics[width=\textwidth]{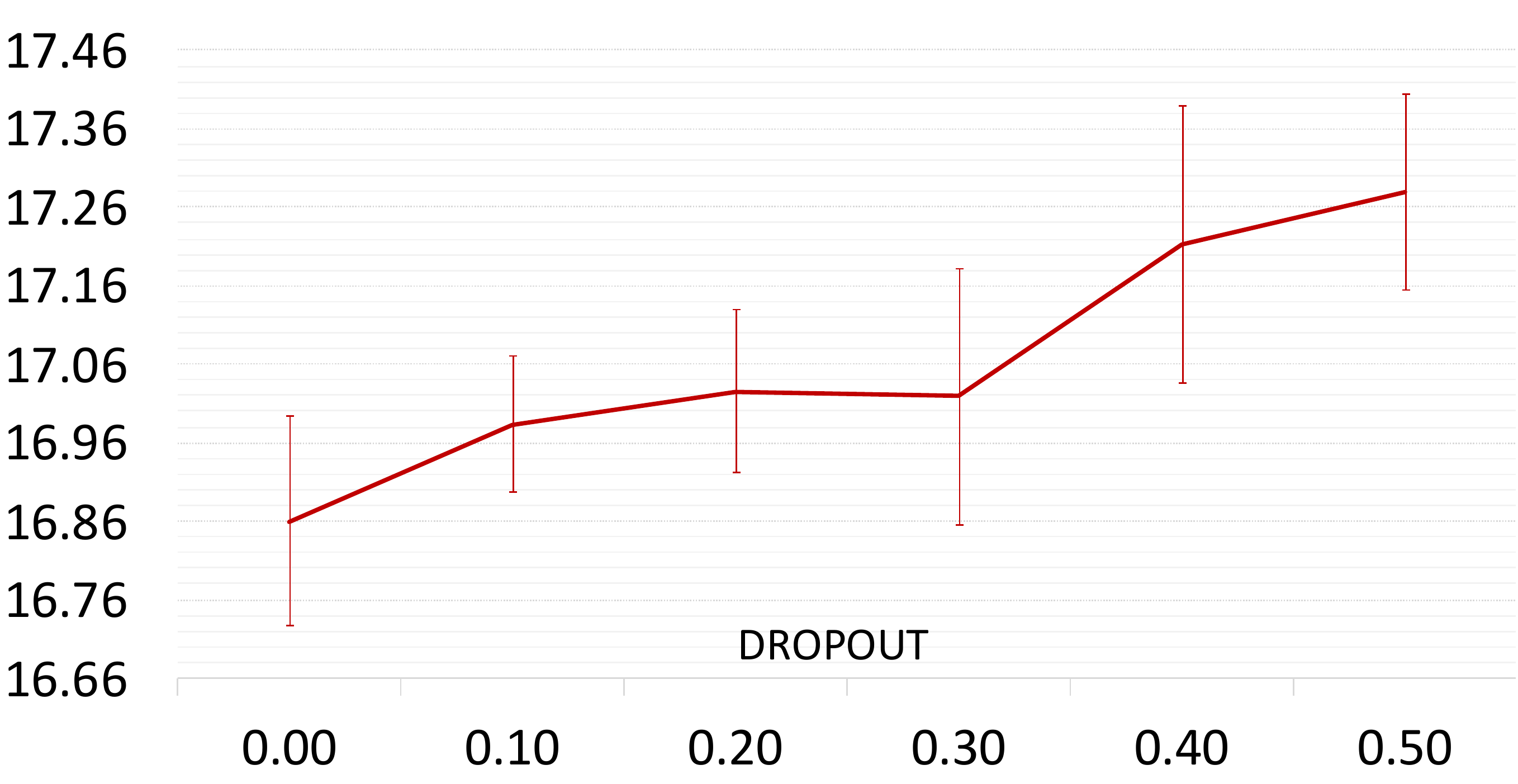}
            \caption[Dropout rate]%
            {{\small Dropout rate}}    
            \label{fig:7d}
        \end{subfigure}
        \hfill
        \begin{subfigure}[b]{0.3\textwidth}   
            \centering 
            \includegraphics[width=\textwidth]{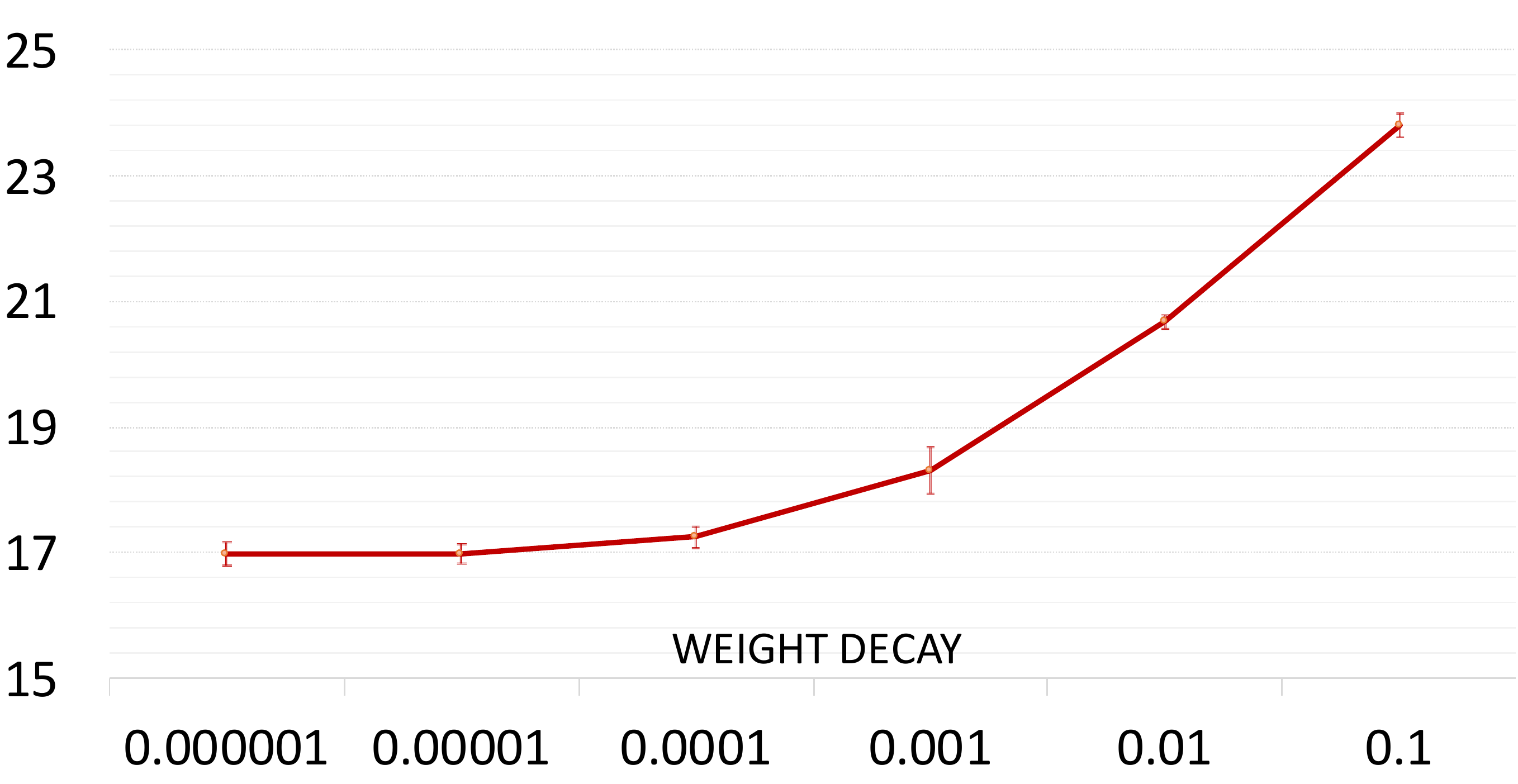}
            \caption[Weight decay]%
            {{\small Weight decay}}    
            \label{fig:7f}
        \end{subfigure}
        \hfill
        \begin{subfigure}[b]{0.3\textwidth}   
            \centering 
            \includegraphics[width=\textwidth]{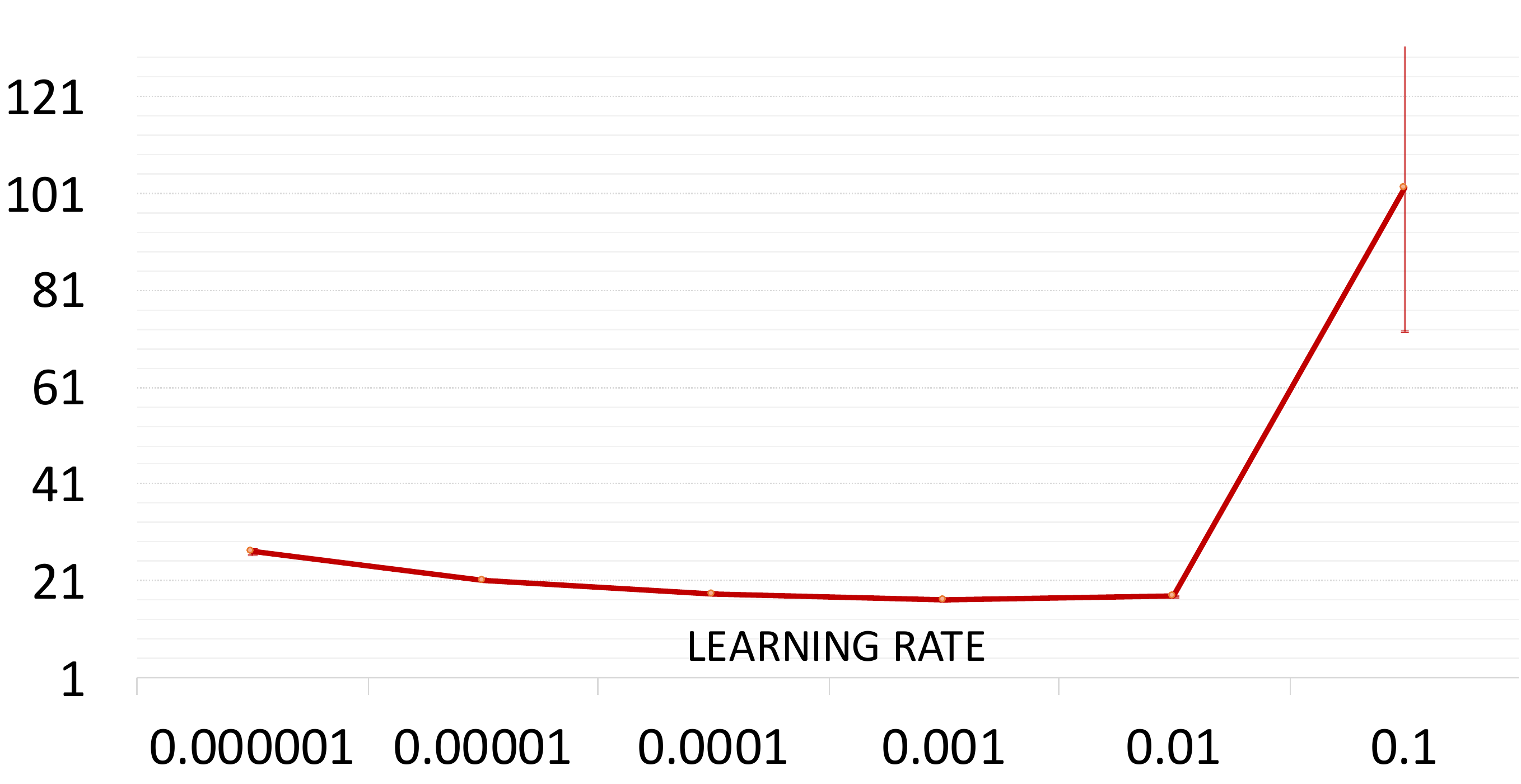}
            \caption[Learning rate]%
            {{\small Learning rate}}    
            \label{fig:7e}
        \end{subfigure}
        
        \caption[Parameter Study]
        {\small MAE plots for hyperparameter study. \textcolor{black}{The y-axis is the MAE score.} The mean with one standard deviation is reported as the value of a hyper-parameter is increased. } 
        \label{fig:para}
\end{figure*}

\subsection{Hyperparameter Study}
To get an understanding of the effect of key hyper-parameters in TraverseNet, we study the effect of varying one hyperparameter at a time whilst keeping others the same as Section~\ref{sec:exptsetting} except that the default number of layers is set to 1 on \textcolor{black}{the} validation set of PEMS-08.
We vary the number of layers ranging from 1 to 6 by 1, the hidden feature dimension ranging from 32 to 192 by 32, the window size ranging from 2 to 12 by 2, the dropout rate ranging from 0 to 0.5 by 0.1, and both the learning rate and weight decay among $\{1\times 10^{-6}, 1\times 10^{-5}, 1\times 10^{-4}, 1\times 10^{-3}, 1\times 10^{-2}, 1\times 10^{-1}\}$. We run each experiment 5 times.  The mean MAE with one standard deviation for each experiment is calculated. Figure \ref{fig:para} plots the trends of model performance for each hyper-parameter. As the number of layers or the window size increases, the model performance is gradually improved. Increasing the number of layers or the window size enlarges the receptive field of a node, thus a node can track longer and broader neighborhood history. The model performance is not sensitive to the change of hidden dimension. We think it may be due to the nature of time series data that the input dimension is thin so a small hidden feature dimension is enough to capture original information.
Dropout and weight decay are not necessary in our model since the model performance drops evidently when it increases. Besides, we observe the optimal learning rate is 0.01.

\subsection{Case Study}
\begin{figure}
\centering 
            \includegraphics[width=0.49\textwidth]{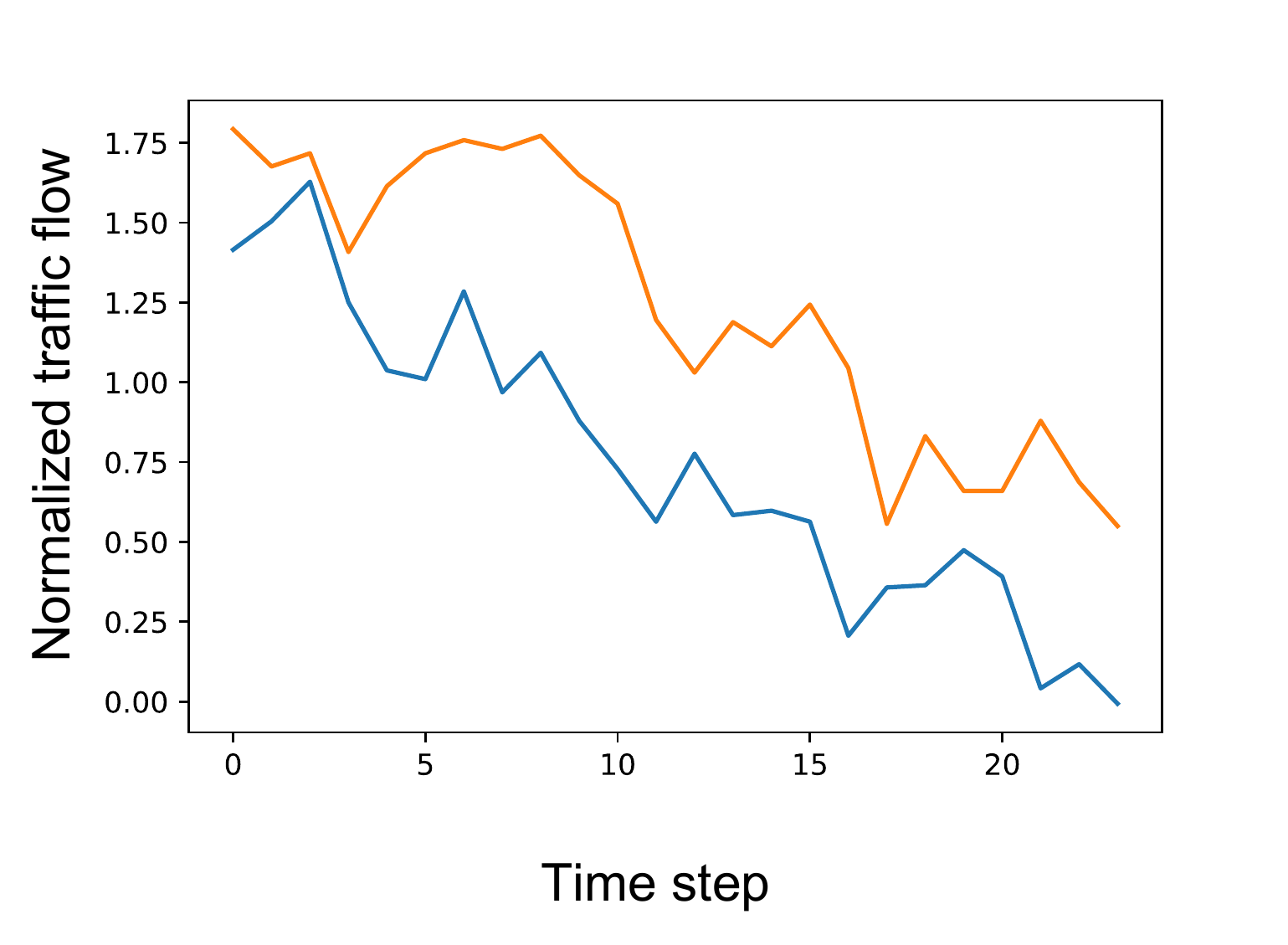}
            \caption{\small\rm Time series of two connected nodes that contain inner spatial-temporal dependencies. The blue line denotes the source node 15. The yellow line denotes the target node 151.}
            \label{fig:casea}
\end{figure}

\begin{figure}
\centering 
            \includegraphics[width=0.49\textwidth]{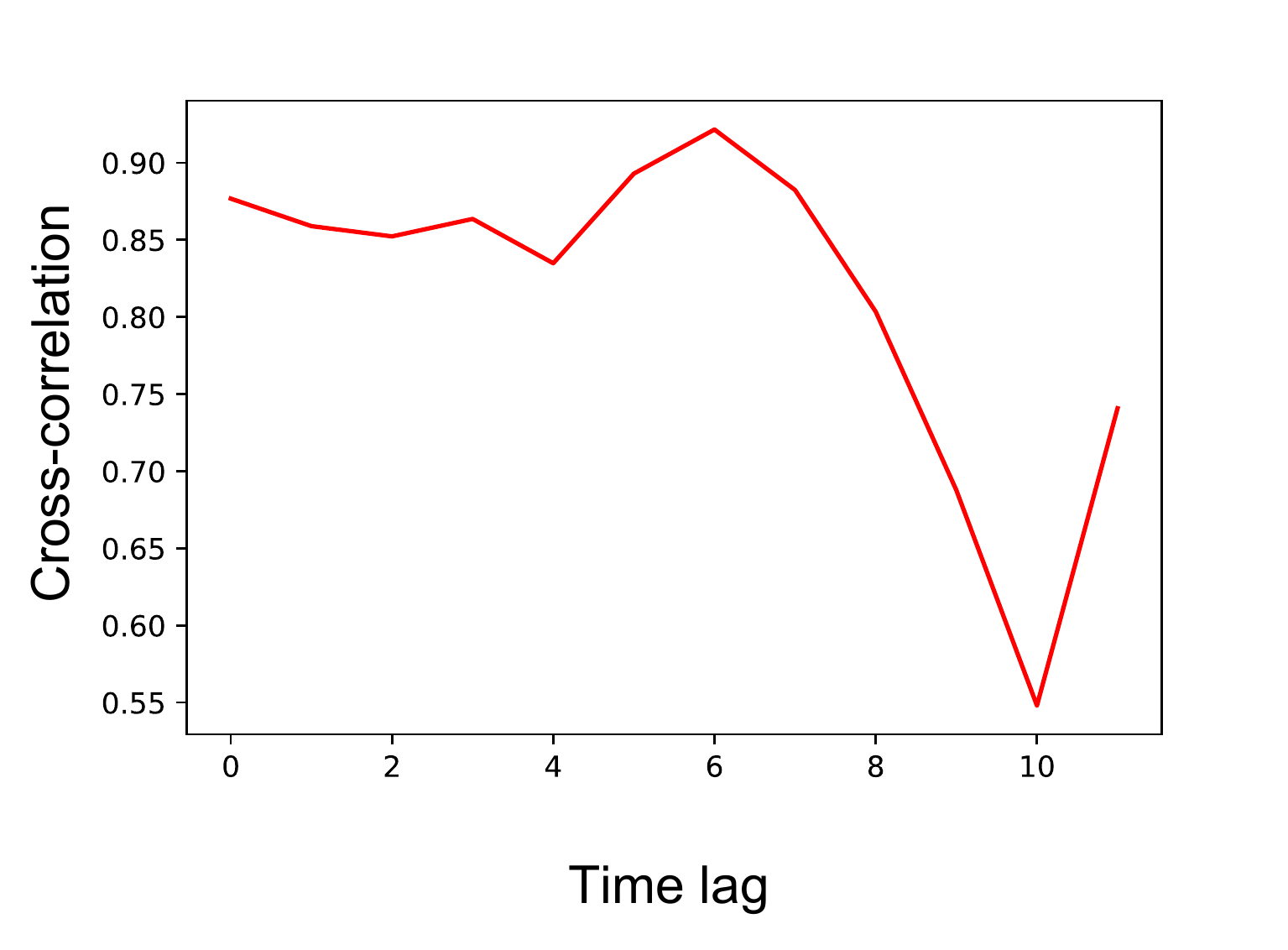}
            \caption{\small\rm Time lag v.s. cross-correlation of the two time series in Fig \ref{fig:casea}. The cross-correlation value peaks at time step 6.}   
            \label{fig:caseb}
\end{figure}

\begin{figure}
            \centering 
            \includegraphics[width=0.49\textwidth]{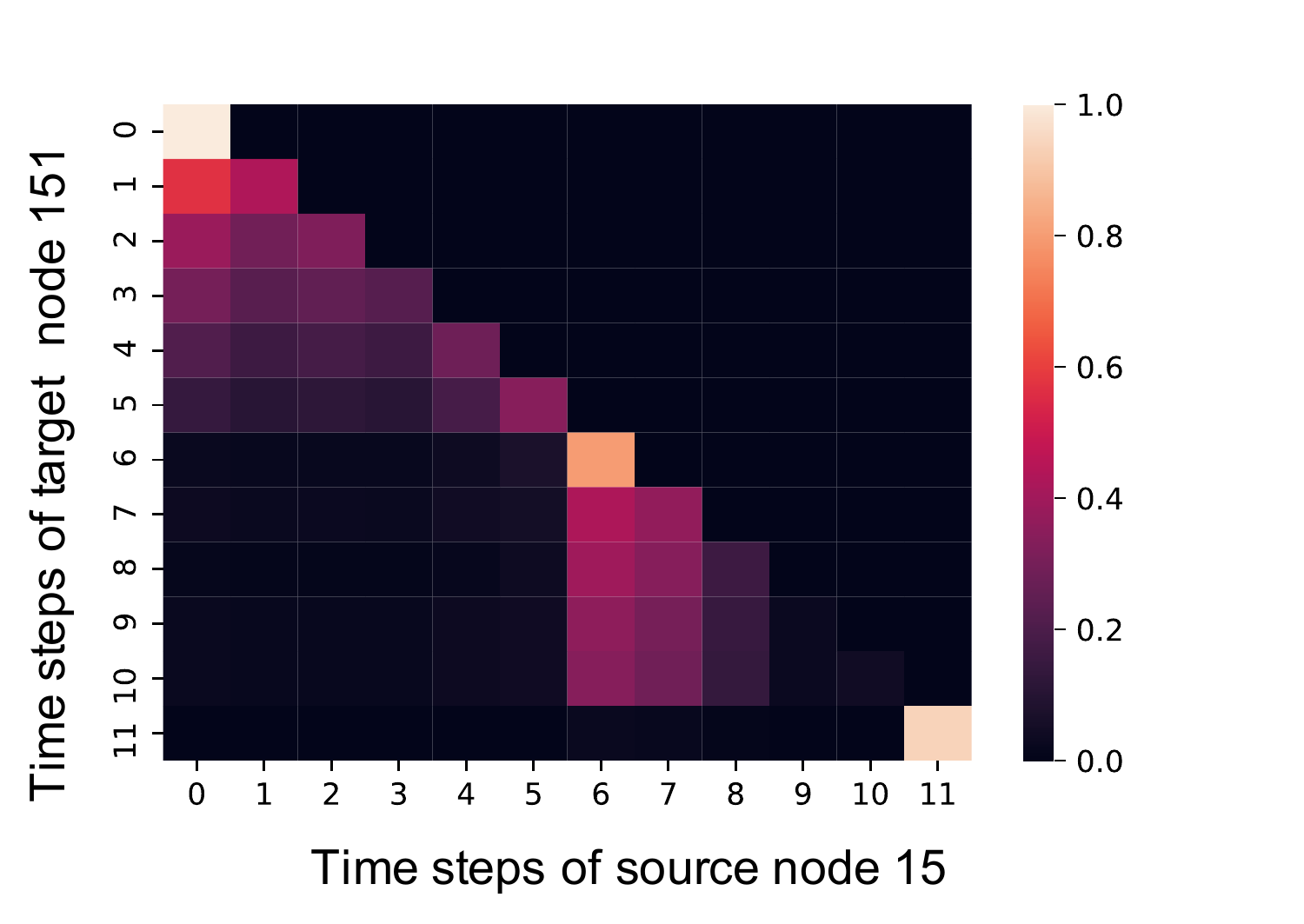}
            \caption{\small\rm Heatmap of attention scores between node 15 and node 151.}    
            \label{fig:casec}
\end{figure}

We perform a case study to understand the effect of TraverseNet in capturing inner spatial-temporal dependency.  Figure \ref{fig:casea} plots the time series of two neighboring nodes, node 15 and node 151 in PEMS-08. The blue line is the time series of the source node 15 and the yellow line is the time series of the target node 151. We observe that the trend of node 151 follows the trend of node 15 with some  extent of latency. For example, the blue line of node 15 starts to drop sharply at step 2 while this phenomenon happens on the yellow line of node 151 6 steps later. Figure \ref{fig:caseb} shows that it is not always the case that the cross-correlation between time series of two neighboring nodes is the highest at time step 0. In fact, 
the time series of node 151 is mostly correlated with the time series of node 15 shifted 6 time steps. Figure \ref{fig:casec} provides a heat-map that visualizes the attention scores produced by TraverseNet in the first message traverse layer between these two time series from time step 0 to time 12.  It shows that the state of node 15 at time step 6, 7, and 8 is very important to the state of node 151 at time step 8, 9 , and 10. This is consistent with the fact the trend of node 15 at time step 6, 7, 8 is similar to the trend of node 151 at time 8, 9, 10 from Figure \ref{fig:casea}. %to do: write a conclusion sentence here.


\subsection{Computation Time}
We compare the computation time of our method with baseline methods on PEMS04 and PEMS08 data in Table \ref{table:computation}. The training speed of DCRNN is the slowest while the training speed of STGCN is the fastest.  The running speed of our method stays in the middle. Although the time complexity of our message traversing layer is $O(M\times p \times Q)$, the speed of our model is still affordable due to an efficient sparse implementation empowered by DGL.

\begin{table}[tb]
		\centering
%Experiments are repeated for two times and we report the mean and standard deviation.}
		%\vskip -0.1in
		
		\caption{Comparison of running time.} 
		\label{table:computation}
		
		\resizebox{0.49\textwidth}{!}{
		\begin{tabular}{l r r |r r}
			\toprule
			\multirow{2}{*}{Models} & \multicolumn{2}{c}{PEMS-04}  & \multicolumn{2}{c}{PEMS-08} \\
			\cline{2-3}  \cline{4-5}
			&\multicolumn{1}{c}{\small Training}  &\multicolumn{1}{c|}{\small Inference} 
			&\multicolumn{1}{c}{\small Training} &\multicolumn{1}{c}{\small Inference}\\
			\midrule
            DCRNN & 314.08 s/epoch & 52.35 s & 145.40 s/epoch & 24.19 s   \\
            STGCN & 10.89 s/epoch & 0.73 s & 6.96 s/epoch & 0.45 s \\
            ASTGCN & 24.11 s/epoch & 2.89 s & 16.41 s/epoch & 1.89 s  \\
            Graph WaveNet & 26.39 s/epoch & 1.59 s & 16.94 s/epoch & 0.91 s \\
            STSGCN & 66.12 s/epoch & 6.01 s & 36.79 s/epoch & 3.27 s \\
            STFGNN & 45.03 s/epoch & 5.19 s & 24.17 s/epoch & 2.82 s \\
            \textbf{TraverseNet} & 56.51 s/epoch & 5.93 s & 39.57 s/epoch & 4.49 s\\
			\bottomrule
		\end{tabular}
		}
	\end{table}

\section{Conclusions}
%In this paper, we propose TraverseNet, a graph neural network that unifies space and time.  The proposed TraverseNet processes a spatial-temporal graph as an inseparable whole.  Through our novel message traverse layers, information can be delivered from the neighbors’ past to the node’s present directly.  Experimental results validate the effectiveness of our framework. There exists several approaches that study graph structure learning in spatial-temporal modeling \cite{wu2020connecting,bai2020adaptive}. A more accurate graph structure may help improve model performance. In future work, we will incorporate graph structure learning into TraverseNet.

In this paper, we propose TraverseNet, a graph neural network that unifies space and time.  The proposed TraverseNet processes a spatial-temporal graph as an inseparable whole.  Through our novel message traverse layers, information can be delivered from the neighbors’ past to the node’s present directly. This design shortens the path length of message passing and enables a node to be aware of its neighborhood variation at firsthand. Experimental results validate the effectiveness of our framework. As the graph structure is a determinant of a spatial-temporal forecasting model \cite{wu2020connecting}, we will consider improving the efficiency of the message traversing by refining the underlying explicit edge relationships in the future.
%To our best knowledge, it is the first time to reveal as well as validate the significance of space-induced temporal delay in spatial-temporal data modeling. Also, the space-induced temporal delay is exploited by traversing information of the neighbors' past to the node's present via our proposed message traverse layers. The phenomenon of space-induced temporal delay discloses the inalienability between spatial correlations and temporal dependency, meanwhile, the proposed TraverseNet processes a spatial-temporal graph as an inseparable whole. Experiments with ablation study and case study validate the importance of the space-induced temporal delay in spatial-temporal data modeling and the effectiveness of the proposed Message Traverse Layer.
%Our  model outperforms 

\bibliographystyle{IEEEtran}
\bibliography{IEEEabrv,biblio}

\begin{thebibliography}{10}
\providecommand{\url}[1]{#1}
\csname url@samestyle\endcsname
\providecommand{\newblock}{\relax}
\providecommand{\bibinfo}[2]{#2}
\providecommand{\BIBentrySTDinterwordspacing}{\spaceskip=0pt\relax}
\providecommand{\BIBentryALTinterwordstretchfactor}{4}
\providecommand{\BIBentryALTinterwordspacing}{\spaceskip=\fontdimen2\font plus
\BIBentryALTinterwordstretchfactor\fontdimen3\font minus
  \fontdimen4\font\relax}
\providecommand{\BIBforeignlanguage}[2]{{%
\expandafter\ifx\csname l@#1\endcsname\relax
\typeout{** WARNING: IEEEtran.bst: No hyphenation pattern has been}%
\typeout{** loaded for the language `#1'. Using the pattern for}%
\typeout{** the default language instead.}%
\else
\language=\csname l@#1\endcsname
\fi
#2}}
\providecommand{\BIBdecl}{\relax}
\BIBdecl

\bibitem{seo2018structured}
Y.~Seo, M.~Defferrard, P.~Vandergheynst, and X.~Bresson, ``Structured sequence
  modeling with graph convolutional recurrent networks,'' in \emph{Advances in
  Neural Information Processing Systems}.\hskip 1em plus 0.5em minus
  0.4em\relax Springer, 2018, pp. 362--373.

\bibitem{li2018diffusion}
Y.~Li, R.~Yu, C.~Shahabi, and Y.~Liu, ``Diffusion convolutional recurrent
  neural network: Data-driven traffic forecasting,'' in \emph{Proceedings of
  the International Conference on Learning Representations}, 2018.

\bibitem{zhang2018gaan}
J.~Zhang, X.~Shi, J.~Xie, H.~Ma, I.~King, and D.-Y. Yeung, ``Gaan: Gated
  attention networks for learning on large and spatiotemporal graphs,'' in
  \emph{Proceedings of the Conference on Uncertainty in Artificial
  Intelligence}, 2018.

\bibitem{yan2018spatial}
S.~Yan, Y.~Xiong, and D.~Lin, ``Spatial temporal graph convolutional networks
  for skeleton-based action recognition,'' in \emph{Proceedings of the AAAI
  conference on artificial intelligence}, vol.~32, no.~1, 2018.

\bibitem{yu2018spatio}
B.~Yu, H.~Yin, and Z.~Zhu, ``Spatio-temporal graph convolutional networks: A
  deep learning framework for traffic forecasting,'' in \emph{Proceedings of
  the Twenty-Seventh I International Joint Conference on Artificial
  Intelligence}, 2018, pp. 3634--3640.

\bibitem{vaswani2017attention}
A.~Vaswani, N.~Shazeer, N.~Parmar, J.~Uszkoreit, L.~Jones, A.~N. Gomez,
  {\L}.~Kaiser, and I.~Polosukhin, ``Attention is all you need,'' in
  \emph{Advances in Neural Information Processing Systems}, vol.~30, 2017, pp.
  5998--6008.

\bibitem{tang2018self}
G.~Tang, M.~M{\"u}ller, A.~Rios, and R.~Sennrich, ``Why self-attention? a
  targeted evaluation of neural machine translation architectures,'' in
  \emph{Proceedings of the Annual Meeting of the Association for Computational
  Linguistics}, 2018, pp. 4263--4272.

\bibitem{guo2019attention}
S.~Guo, Y.~Lin, N.~Feng, C.~Song, and H.~Wan, ``Attention based
  spatial-temporal graph convolutional networks for traffic flow forecasting,''
  in \emph{Proceedings of AAAI Conference on Artificial Intelligence}, vol.~33,
  2019, pp. 922--929.

\bibitem{park2020stgrat}
C.~Park, C.~Lee, H.~Bahng, K.~Kim, S.~Jin, S.~Ko, J.~Choo \emph{et~al.},
  ``Stgrat: a spatio-temporal graph attention network for traffic
  forecasting,'' in \emph{Proceedings of the Conference on Information and
  Knowledge Management}, 2020.

\bibitem{wang2020traffic}
X.~Wang, Y.~Ma, Y.~Wang, W.~Jin, X.~Wang, J.~Tang, C.~Jia, and J.~Yu, ``Traffic
  flow prediction via spatial temporal graph neural network,'' in
  \emph{Proceedings of the World Wide Web Conference}, 2020, pp. 1082--1092.

\bibitem{kipf2017semi}
T.~N. Kipf and M.~Welling, ``Semi-supervised classification with graph
  convolutional networks,'' in \emph{Proceedings of the International
  Conference on Learning Representations}, 2017.

\bibitem{levie2017cayleynets}
R.~Levie, F.~Monti, X.~Bresson, and M.~M. Bronstein, ``Cayleynets: Graph
  convolutional neural networks with complex rational spectral filters,''
  \emph{IEEE Transactions on Signal Processing}, vol.~67, no.~1, pp. 97--109,
  2018.

\bibitem{bianchi2021graph}
F.~M. Bianchi, D.~Grattarola, L.~Livi, and C.~Alippi, ``Graph neural networks
  with convolutional arma filters,'' \emph{IEEE Transactions on Pattern
  Analysis and Machine Intelligence}, 2021.

\bibitem{hamilton2017inductive}
W.~Hamilton, Z.~Ying, and J.~Leskovec, ``Inductive representation learning on
  large graphs,'' in \emph{Advances in Neural Information Processing Systems},
  2017, pp. 1024--1034.

\bibitem{gilmer2017neural}
J.~Gilmer, S.~S. Schoenholz, P.~F. Riley, O.~Vinyals, and G.~E. Dahl, ``Neural
  message passing for quantum chemistry,'' in \emph{Proc. of ICML}, 2017, pp.
  1263--1272.

\bibitem{wu2020comprehensive}
Z.~Wu, S.~Pan, F.~Chen, G.~Long, C.~Zhang, and S.~Y. Philip, ``A comprehensive
  survey on graph neural networks,'' \emph{IEEE transactions on neural networks
  and learning systems}, 2020.

\bibitem{wan2021contrastive}
S.~Wan, Y.~Zhan, L.~Liu, B.~Yu, S.~Pan, and C.~Gong, ``Contrastive graph
  poisson networks: Semi-supervised learning with extremely limited labels,''
  \emph{Advances in Neural Information Processing Systems}, vol.~34, 2021.

\bibitem{liu2021graph}
Y.~Liu, M.~Jin, S.~Pan, C.~Zhou, Y.~Zheng, F.~Xia, and P.~Yu, ``Graph
  self-supervised learning: A survey,'' \emph{IEEE Transactions on Knowledge
  and Data Engineering}, 2022.

\bibitem{zhang2022trustworthy}
H.~Zhang, B.~Wu, X.~Yuan, S.~Pan, H.~Tong, and J.~Pei, ``Trustworthy graph
  neural networks: Aspects, methods and trends,'' \emph{arXiv:2205.07424},
  2022.

\bibitem{monti2017geometric}
F.~Monti, D.~Boscaini, J.~Masci, E.~Rodola, J.~Svoboda, and M.~M. Bronstein,
  ``Geometric deep learning on graphs and manifolds using mixture model cnns,''
  in \emph{Proceedings of 2017 IEEE Conference on Computer Vision and Pattern
  Recognition}.\hskip 1em plus 0.5em minus 0.4em\relax IEEE, 2017.

\bibitem{velickovic2017graph}
P.~Velickovic, G.~Cucurull, A.~Casanova, A.~Romero, P.~Lio, and Y.~Bengio,
  ``Graph attention networks,'' in \emph{Proceedings of the 5th International
  Conference on Learning Representations}, 2017.

\bibitem{gao2019graph}
H.~Gao and S.~Ji, ``Graph representation learning via hard and channel-wise
  attention networks,'' in \emph{Proceedings of the 25th ACM SIGKDD
  International Conference on Knowledge Discovery \& Data Mining}, 2019, pp.
  741--749.

\bibitem{klicpera2019diffusion}
J.~Klicpera, S.~Wei{\ss}enberger, and S.~G{\"u}nnemann, ``Diffusion improves
  graph learning,'' in \emph{Advances in Neural Information Processing
  Systems}, 2019, pp. 13\,354--13\,366.

\bibitem{wang2020direct}
G.~Wang, R.~Ying, J.~Huang, and J.~Leskovec, ``Direct multi-hop attention based
  graph neural network,'' \emph{arXiv preprint arXiv:2009.14332}, 2020.

\bibitem{jin2022multivariate}
M.~Jin, Y.~Zheng, Y.-F. Li, S.~Chen, B.~Yang, and S.~Pan, ``Multivariate time
  series forecasting with dynamic graph neural {ODEs},'' \emph{arXiv
  2202.08408}, 2022.

\bibitem{pan2019urban}
Z.~Pan, Y.~Liang, W.~Wang, Y.~Yu, Y.~Zheng, and J.~Zhang, ``Urban traffic
  prediction from spatio-temporal data using deep meta learning,'' in
  \emph{KDD}.\hskip 1em plus 0.5em minus 0.4em\relax ACM, 2019, pp. 1720--1730.

\bibitem{li2019spatio}
B.~Li, X.~Li, Z.~Zhang, and F.~Wu, ``Spatio-temporal graph routing for
  skeleton-based action recognition,'' in \emph{Proceedings of AAAI Conference
  on Artificial Intelligence}, 2019, pp. 8561--8568.

\bibitem{wu2019graph}
Z.~Wu, S.~Pan, G.~Long, J.~Jiang, and C.~Zhang, ``Graph wavenet for deep
  spatial-temporal graph modeling,'' in \emph{Proceedings of the Twenty-Eighth
  International Joint Conference on Artificial Intelligence}, 2019.

\bibitem{wu2020connecting}
Z.~Wu, S.~Pan, G.~Long, J.~Jiang, X.~Chang, and C.~Zhang, ``Connecting the
  dots: Multivariate time series forecasting with graph neural networks,'' in
  \emph{Proceedings of the 26th ACM SIGKDD International Conference on
  Knowledge Discovery \& Data Mining}, 2020, pp. 753--763.

\bibitem{zhang2020spatio}
Q.~Zhang, J.~Chang, G.~Meng, S.~Xiang, and C.~Pan, ``Spatio-temporal graph
  structure learning for traffic forecasting,'' in \emph{Proceedings of the
  AAAI Conference on Artificial Intelligence}, vol.~34, no.~01, 2020, pp.
  1177--1185.

\bibitem{zheng2020gman}
C.~Zheng, X.~Fan, C.~Wang, and J.~Qi, ``Gman: A graph multi-attention network
  for traffic prediction,'' in \emph{Proceedings of AAAI Conference on
  Artificial Intelligence}, 2020.

\bibitem{song2020spatial}
C.~Song, Y.~Lin, S.~Guo, and H.~Wan, ``Spatial-temporal synchronous graph
  convolutional networks: A new framework for spatial-temporal network data
  forecasting,'' in \emph{Proceedings of AAAI Conference on Artificial
  Intelligence}, vol.~34, no.~01, 2020, pp. 914--921.

\bibitem{mengzhang2020spatial}
L.~Mengzhang and Z.~Zhanxing, ``Spatial-temporal fusion graph neural networks
  for traffic flow forecasting,'' in \emph{Proceedings of AAAI Conference on
  Artificial Intelligence}, 2021.

\bibitem{pan2020spatio}
C.~Pan, S.~Chen, and A.~Ortega, ``Spatio-temporal graph scattering transform,''
  in \emph{International Conference on Learning Representations}, 2021.

\bibitem{hadou2021space}
S.~Hadou, C.~I. Kanatsoulis, and A.~Ribeiro, ``Space-time graph neural
  networks,'' in \emph{International Conference on Learning Representations},
  2022.

\bibitem{isufi2021graph}
E.~Isufi and G.~Mazzola, ``Graph-time convolutional neural networks,'' in
  \emph{2021 IEEE Data Science and Learning Workshop (DSLW)}.\hskip 1em plus
  0.5em minus 0.4em\relax IEEE, 2021, pp. 1--6.

\bibitem{zambon2019autoregressive}
D.~Zambon, D.~Grattarola, L.~Livi, and C.~Alippi, ``Autoregressive models for
  sequences of graphs,'' in \emph{2019 International Joint Conference on Neural
  Networks (IJCNN)}.\hskip 1em plus 0.5em minus 0.4em\relax IEEE, 2019, pp.
  1--8.

\bibitem{paassen2020graph}
B.~Paassen, D.~Grattarola, D.~Zambon, C.~Alippi, and B.~E. Hammer, ``Graph edit
  networks,'' in \emph{International Conference on Learning Representations},
  2020.

\bibitem{wang2019dgl}
M.~Wang, D.~Zheng, Z.~Ye, Q.~Gan, M.~Li, X.~Song, J.~Zhou, C.~Ma, L.~Yu,
  Y.~Gai, T.~Xiao, T.~He, G.~Karypis, J.~Li, and Z.~Zhang, ``Deep graph
  library: A graph-centric, highly-performant package for graph neural
  networks,'' \emph{arXiv preprint arXiv:1909.01315}, 2019.

\bibitem{sutskever2014sequence}
I.~Sutskever, O.~Vinyals, and Q.~V. Le, ``Sequence to sequence learning with
  neural networks,'' \emph{arXiv preprint arXiv:1409.3215}, 2014.

\bibitem{cho2014learning}
K.~Cho, B.~van Merri{\"e}nboer, C.~Gulcehre, D.~Bahdanau, F.~Bougares,
  H.~Schwenk, and Y.~Bengio, ``Learning phrase representations using {RNN}
  encoder{--}decoder for statistical machine translation,'' in
  \emph{Proceedings of the 2014 Conference on Empirical Methods in Natural
  Language Processing ({EMNLP})}, 2014, pp. 1724--1734.

\end{thebibliography}
%\newpage
%\input{bio.tex}
\balance

\end{document}